\documentclass[sigconf]{acmart}

\usepackage[utf8]{inputenc} 

\newcommand{\cev}[1]{\reflectbox{\ensuremath{\vec{\reflectbox{\ensuremath{#1}}}}}}

\setcopyright{iw3c2w3}
\copyrightyear{2020}
\acmYear{2020}
\acmDOI{10.1145/3366423.3380208}

\acmConference[WWW '20]{Proceedings of The Web Conference 2020}{April 20--24, 2020}{Taipei, Taiwan}
\acmBooktitle{Proceedings of The Web Conference 2020 (WWW '20), April 20--24, 2020, Taipei, Taiwan} 
\acmPrice{}
\acmISBN{978-1-4503-7023-3/20/04}

\acmSubmissionID{1201}

\begin{document}


\title{Learning Contextualized Document Representations for Healthcare Answer Retrieval}


\author{Sebastian Arnold}
\email{sarnold@beuth-hochschule.de}
\orcid{0000-0001-7788-6100}
\affiliation{%
  \institution{Beuth University of Applied Sciences}
  \streetaddress{Luxemburger Strasse 10}
  \postcode{13353}
  \city{Berlin}
  \country{Germany}}
\author{Betty van Aken}
\email{bvanaken@beuth-hochschule.de}
\affiliation{%
  \institution{Beuth University of Applied Sciences}
  \streetaddress{Luxemburger Strasse 10}
  \postcode{13353}
  \city{Berlin}
  \country{Germany}}
\author{Paul Grundmann}
\email{s76413@beuth-hochschule.de}
\affiliation{%
  \institution{Beuth University of Applied Sciences}
  \streetaddress{Luxemburger Strasse 10}
  \postcode{13353}
  \city{Berlin}
  \country{Germany}}
\author{Felix A. Gers}
\email{gers@beuth-hochschule.de}
\affiliation{%
  \institution{Beuth University of Applied Sciences}
  \streetaddress{Luxemburger Strasse 10}
  \postcode{13353}
  \city{Berlin}
  \country{Germany}}
\author{Alexander L\"oser}
\email{aloeser@beuth-hochschule.de}
\affiliation{%
  \institution{Beuth University of Applied Sciences}
  \streetaddress{Luxemburger Strasse 10}
  \postcode{13353}
  \city{Berlin}
  \country{Germany}}

\renewcommand{\shortauthors}{Arnold, van Aken, Grundmann, Gers and L\"oser}

\begin{abstract}
We present Contextual Discourse Vectors (CDV), a distributed document representation for efficient answer retrieval from long healthcare documents. Our approach is based on structured query tuples of entities and aspects from free text and medical taxonomies. Our model leverages a dual encoder architecture with hierarchical LSTM layers and multi-task training to encode the position of clinical entities and aspects alongside the document discourse. We use our continuous representations to resolve queries with short latency using approximate nearest neighbor search on sentence level. We apply the CDV model for retrieving coherent answer passages from nine English public health resources from the Web, addressing both patients and medical professionals. Because there is no end-to-end training data available for all application scenarios, we train our model with self-supervised data from Wikipedia. We show that our generalized model significantly outperforms several state-of-the-art baselines for healthcare passage ranking and is able to adapt to heterogeneous domains without additional fine-tuning.
\end{abstract}




\maketitle

\section{Introduction}
\label{sec:introduction}

\begin{figure}[t]
\includegraphics[]{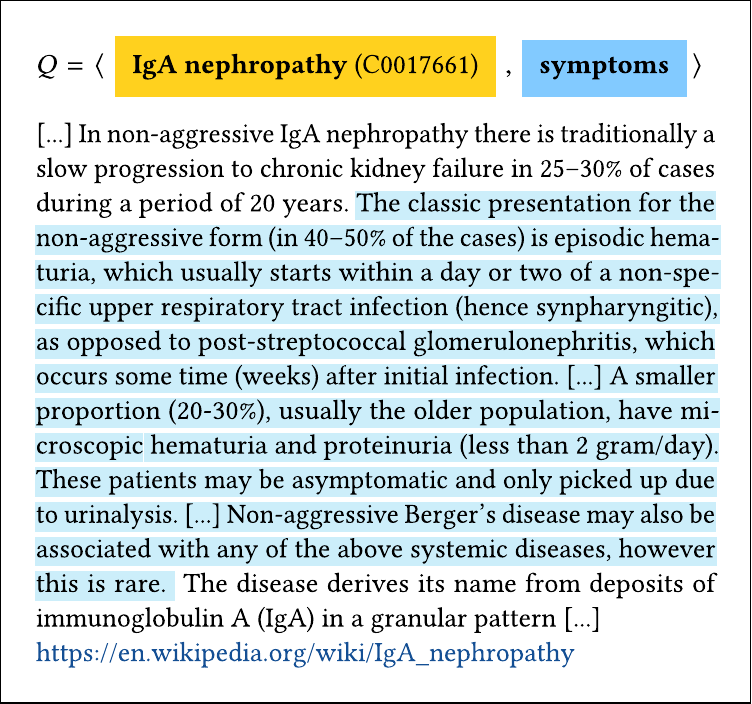}
\Description[The query Q=(IgA nephropathy, symptoms) and a snippet of text from Wikipedia.]{The query Q=(IgA nephropathy, symptoms) and a snippet of text taken from the Wikipedia article about IgA nephropathy, where a long answer passage is highlighted.}
\caption{Example of a structured entity/aspect query $Q$ and a highlighted answer passage from Wikipedia. Note that the answer is part of a longer document and there is almost no word overlap between query and answer passage.}
\label{fig:example}
\end{figure}

In a clinical decision support system (CDSS), doctors and healthcare professionals require access to information from heterogeneous sources, such as research papers \cite{gorman1994can, schardt2007utilization}, electronic health records \cite{hanauer2015supporting}, clinical case reports \cite{fujiwara2018pubcasefinder}, reference works and knowledge base articles. Differential diagnosis is an important task where a doctor seeks to retrieve answers for non-factoid queries about diseases, such as \textsl{``symptoms of IgA nephropathy''} (see Figure \ref{fig:example}). A relevant answer typically spans multiple sentences and is most likely embedded into the discourse of a long document \cite{yang2016beyond, cohan2018discourse}. 

Evidence-based medicine (EBM) has made efforts to structure physicians' information needs into short structured question representations, such as PICO (patient, intervention, comparison, outcome) \cite{richardson1995well} and---more general---well-formed background / foreground questions \cite{cheng2004study}. We support this important query intention and define a query as structured tuple of \emph{entity} (e.g. a disease or health problem) and \emph{aspect}. Our model is focused on clinical aspects such as therapy, diagnosis, etiology, prognosis, and others, which have been described in the literature previously by manual clustering of semantic question types \cite{huang2006evaluation} or crawling medical Wikipedia section headings \cite{arnold2019sector}. In a CDSS, a doctor can express these query terms with identifiers from a knowledge base or medical taxonomy, e.g. UMLS, ICD-10 or Wikidata. The system will support the user in assigning these links by search and auto-completion operators \cite{schneider2018smartmd, fujiwara2018pubcasefinder}, which allows us to use these representations as input for the answer retrieval task. 

Several methods have been proposed to apply deep neural networks for effective information retrieval \cite{guo2016deep, mitra2017learning, dai2018convolutional} and question answering \cite{seo2017bidirectional, wang2017gated}, also with focus on healthcare \cite{zhu2019hierarchical, jin2019pubmedqa}. However, our CDSS scenario poses a unique combination of open challenges to a retrieval system:

\begin{enumerate}

\item \emph{Task coverage:} Query intentions span a broad range in specificy and complexity \cite{huang2006evaluation, nanni2017benchmark}. For example, medical specialists may pose very precise queries that align with a pre-defined taxonomy and focus on rare diseases. On the other hand, nursing staff might have broader and more heterogeneous questions. However, in most cases we do not have access to task-specific training data, so training the model for a single intention is not feasible. We therefore require a generalized query representation that covers a broad range of intents and taxonomies, even with limited training data.

\item \emph{Domain adaptability:} In many cases we do not even have textual data readily available at training time from all resources in a CDSS. However, we observe linguistic and semantic shifts between the heterogeneous types of text, e.g. different use of terms and abbreviations among groups of doctors. Therefore we face a zero-shot retrieval task that requires robust domain transfer abilities across diverse biomedical, clinical and healthcare text resources \cite{logeswaran2019zero}.

\item \emph{Contextual coherence:} Answers are often expressed as passages in context of a long document. Therefore the model needs to respect long-range dependencies such as the sequence of micro-topics that establish a coherent `train of thought' in a document \cite{arora2016latent, arnold2019sector}. At the same time, the model is required to operate on a fine granularity (e.g., on sentence level) rather than on entire documents to be able to capture the boundaries of answers \cite{keikha2014retrieving}.

\item \emph{Efficient neural information retrieval:} Finally, all documents in the CDSS need to be accessible with fast ad-hoc queries by the users. Many question answering models are based on pairwise similarity, which is computationally too intensive when applied to large-scale retrieval tasks \cite{gillick2018end}. Instead, we require a continuous retrieval model that allows for offline indexing and approximate nearest neighbor search with high recall \cite{gillick2018end}, even for rare queries and with low latency in the order of milliseconds.

\end{enumerate}


We approach these challenges and present \emph{Contextual Discourse Vectors} (CDV)\footnote{Code and evaluation data is available at \url{https://github.com/sebastianarnold/cdv}}, a neural document representation which is based on discourse modeling and fulfills the above requirements. Our method is the first to address answer retrieval with structured queries on long heterogeneous documents from the healthcare domain. 

CDV is based on hierarchical layers to encode word, sentence and document context with bidirectional Long Short-Term Memory (BLSTM). The model uses multi-task learning \cite{caruana1997multitask} to align the sequence of sentences in a long document to the clinical knowledge encoded in pre-trained entity and aspect vector spaces. We use a dual encoder architecture \cite{gillick2018end}, which allows us to precompute discourse vectors for all documents and later answer ad-hoc queries over that corpus with short latency \cite{gillick2019learning}. Consequently, the model predicts similarity scores with sentence granularity and does not require an extra inference step after the initial document indexing.

We apply our CDV model for retrieving passages from various public health resources on the Web, including NIH documents and Patient articles, with structured clinical query intentions of the form $\langle\text{entity}, \text{aspect}\rangle$. Because there is no training data available from most sources, we use a self-supervised approach to train a generalized model from medical Wikipedia texts. We apply this model to the texts in our evaluation in a zero-shot approach \cite{palatucci2009zero} without additional fine tuning. 

In summary, the major contributions of this paper include:
\begin{itemize}

\item We propose a structured entity/aspect healthcare query model to support the essential query intentions of medical professionals. Our task is focused on the efficient retrieval of answer passages from long documents of heterogeneous health resources. 

\item We introduce CDV, a contextualized document representation for passage retrieval. Our model leverages a dual encoder architecture with BLSTM layers and multi-task training to encode the position of discourse topics alongside the document. We use the representations to answer queries using nearest neighbor search on sentence level.

\item Our model utilizes generalized language models and aligns them with clinical knowledge from medical taxonomies, e.g. pre-trained entity and aspect embeddings. Therefore, it can be trained with sparse self-supervised training data, e.g. from Wikipedia texts, and is applicable to a broad range of texts.

\item We prove the applicability of our CDV model with extensive experiments and a qualitative error analysis on nine heterogeneous healthcare resources. We provide additional entity/aspect labels for all datasets. Our model significantly outperforms existing document matching methods in the retrieval task and can adapt to different healthcare domains without fine-tuning.

\end{itemize}

In this paper, we first give an overview of related research (Section \ref{sec:background}). Next, we introduce our query representation (Section \ref{sec:discourse}). Then, we focus on the contextual document representation model (Section \ref{sec:model}). Finally, we discuss the findings of our experimental evaluation of the healthcare retrieval task (Section \ref{sec:evaluation}) and summarize our conclusions (Section \ref{sec:conclusion}).

\section{Background}
\label{sec:background}

There is a large amount of work on \emph{question answering} (QA) \cite{seo2017bidirectional, wang2017gated}, also applied to healthcare \cite{abacha2019overview, jin2019pubmedqa} which focuses primarily on factoid questions with short answers. Typically, these models are trained with labeled question-answer pairs. However, it was shown that these models are not suitable for extracting local aspects from long documents, and especially not for open-ended, long answer passages \cite{tellex2003quantitative, keikha2014retrieving, yang2016beyond, zhu2019hierarchical}. We therefore frame our task as a \emph{passage retrieval} problem, where the system's goal is to extract a concise snippet (typically 5--20 sentences) out of a large number of long documents. Furthermore, following studies from EBM \cite{richardson1995well, cheng2004study, huang2006evaluation}, we focus on \emph{structured healthcare queries} instead of free-text questions.

\paragraph{Discourse-aware representations}

Recently, new approaches have emerged that represent local information in the context of long documents. 
For example, \citet{cohan2018discourse} approach the problem as abstractive summarization task. The authors use hierarchical encoders to model the discourse structure of a document and generate summaries using an attentive discourse-aware decoder. 
In our prior work on SECTOR \cite{arnold2019sector}, we apply a segmentation and classification method to long documents to identify coherent passages and classify them into 27 clinical aspects. The model produces a continuous topic embedding on sentence level using BLSTMs, which has similar properties to the micro-topics described earlier by \citet{arora2016latent} as \emph{discourse vector} (``what is being talked about'').

We follow these ideas as the groundwork for our approach. Our proposed model is based on a hierarchical architecture to encode a continuous discourse representation. To the best of our knowledge, our model is the first to use discourse-aware representations for answer retrieval. Additionally, we address the problem of sparse training data and propose a multi-task approach for training the model with self-supervised data instead of labeled examples.

\paragraph{Passage matching}

A baseline approach to the passage retrieval problem is to split longer documents into individual passages and rank them independently according to their relevance for the query. Passage matching has been done using term-based methods \cite{robertson1976relevance, salton1988term}, most prominently in TF-IDF \cite{sparck1972statistical} or Okapi BM25 \cite{robertson1995okapi}. However, these methods usually do not perform well on long passages or when there is minimal word overlap between passage and query. Therefore, most neural matching models tackle this vocabulary mismatch using semantic vector-space representations.

Representation-based matching models aim to match the continuous representations of queries and passages using a similarity function, e.g. cosine distance. This can be done on sentence level (ARC-I \cite{hu2014convolutional}), which does not work well if queries are short and passages are longer than a few sentences. Therefore, most approaches learn distinct query and passage representations using feed-forward (DSSM \cite{huang2013learning}) or CNN convolutional neural networks (C-DSSM \cite{shen2014learning}).

Interaction-based matching models focus on the complex interaction between query and passage. These models use CNNs on sentence level (ARC-II \cite{hu2014convolutional}), match query terms and words using word count histograms (DRMM \cite{guo2016deep}), word-level dot product similarity (MatchPyramid \cite{pang2016text}), attention-based neural networks (aNMM \cite{yang2016anmm}), kernel pooling (K-NRM \cite{xiong2017end}) or convolutional n-gram kernel pooling (Conv-KNRM \cite{dai2018convolutional}). Eventually, \citet{zhu2019hierarchical} utilize hierarchical attention on word and sentence level (HAR) to capture interaction of the query with local context in long passages. 

While interaction-based models can capture complex correlations between query and passage, these models do not include contextualized local information---e.g. long-range document context that comes before or after a passage---which might contain important information for the query. To overcome this problem, \citet{mitra2017learning} combine document-level representations with interaction features in a deep CNN model (Duet). \citet{wan2016deep}  utilize BLSTMs (MVLSTM) to generate positional sentence representations across the entire document.

We combine the representation approach with interaction. Our proposed model is able to learn the interaction between the words of the passage and the discourse using a hierarchical architecture. At the same time, it encodes fixed sentence representations that we use to match query representations. Consequently, our model does not require pairwise inference between all query--sentence pairs, which is usually circumvented by re-ranking candidates \cite{gillick2018end}. Instead, our model requires only a single pass through all documents at index time. Furthermore, by encoding discourse-aware representations, the model is able to access long-range document context which is normally hidden after the passage split. We compare our approach to all the discussed matching models and review these properties again in Section \ref{sec:evaluation}. 

\section{Query Model}
\label{sec:discourse}

Our first challenge is to design a query model which can adapt to a broad number of healthcare answer retrieval tasks and utilizes the information sources available in a CDSS. In this section, we introduce a vector-space representation for this purpose.

We define a query as a structured tuple $Q=\langle\text{entity},\text{aspect}\rangle$. This approach of using two complementary query arguments originates from the idea of structured background/foreground questions in EBM \cite{cheng2004study} and has been used before in many triple-based retrieval systems \cite{adolphs2011yago}.
In our healthcare scenario, we restrict entities to be of type \emph{disease}, e.g. \textsl{``IgA nephropathy''}, and aspects from the clinical domain, e.g. \textsl{``symptoms''}, \textsl{``treatment''}, or \textsl{``prognosis''}. We discuss these two spaces in Sections \ref{sec:discourse_entity} and Section \ref{sec:discourse_aspect} and propose their combination in Section \ref{sec:query_representation}. In general, our model is not limited to the query spaces used in this paper and further extendable to a larger number of arguments. 

\subsection{Entity Space}
\label{sec:discourse_entity}

The first part of our problem is to represent the entity in focus of a query. In contrast to interaction-based models, which are applied to query--document pairs, our approach is to decouple entity encoding and document encoding. Therefore we follow recent work in representation-based Entity Linking \cite{gillick2019learning} and embed textual knowledge from the clinical domain into this representation. Our goal is to generalize entity representations, so the model will be able to align to existing taxonomies without retraining. Therefore, our entity space must be as complete as possible: it needs to cover each of the entities that appear in the discourse training data, but also rare entities that we expect at query time, e.g. in the application. We must further provide a robust method for predicting unseen entities \cite{logeswaran2019zero}. In contrast to highly specialized entity embeddings constructed from knowledge graphs or multimodal data \cite{beam2018clinical}, our generic approach is based on textual data and allows us to apply the model to different knowledge bases and domains.

\subsubsection{Entity Embeddings}
\label{sec:entityembeddings}

Our goal is to create a mapping of each entity in the knowledge base  $E \in \mathcal{K}$ identified by its ID into a low-dimensional entity vector space $\mathbb{E} \subset \mathbb{R}^d$\footnote{we use $d$ as a placeholder for all embedding vector sizes, even if they are not equal}. We train an embedding by minimizing the loss for predicting the entity from sentences $s \in \mathcal{D}_E$ in the entity descriptions:
\begin{eqnarray}
\mathcal{L}_\text{entity}(\Theta) = -\text{log} \prod_{E \in \mathcal{K}} \prod_{s \in \mathcal{D}_E} p_\Theta\bigl(E\text{.id} \mid s\bigr)
\end{eqnarray}
where $\Theta$ denotes the parameters required to approximate the probability $p$. We optimize $\Theta$ using a bidirectional Long Short-Term Memory (BLSTM) \cite{hochreiter1997long} to predict the entity ID $E\text{.id}$ from the words $w_i \in s$. We encode $w_i$ using Fasttext embeddings \cite{bojanowski2017enriching} and use bloom filters \cite{serra2017getting} to compress $E\text{.id}$ into a hashed bit encoding, allowing for less model parameters and faster training.
\begin{equation}
\begin{alignedat}{2}
p_\Theta\bigl(E\text{.id} \mid s\bigr) &= p_\Theta\bigl(E\text{.id} \mid w_1, \ldots, w_N\bigr) \\
&\approx p\bigl(\text{bloom}(E\text{.id}) \mid \text{BLSTM}_\Theta(w_1, \ldots, w_N)\bigr)
\end{alignedat}
\end{equation}

Subsequently, we extend the approach of \citet{palangi2016deep} and define the embedding function $\epsilon$ as the average output of the hidden word states $\vec{g}_k$ and $\cev{g}_k$ at the first respectively last time step:
\begin{equation}
\label{eq:blstm}
\begin{alignedat}{2}
\vec{g}_k &= \text{LSTM}_\Theta(\vec{g}_{k-1}, w_k) \\
\cev{g}_k &= \text{LSTM}_\Theta(\cev{g}_{k+1}, w_k)
\end{alignedat}
\end{equation}
\begin{equation}
\label{eq:semb}
\epsilon(s) = \frac{\vec{g}_{T} + \cev{g}_{1}}{2}
\end{equation}

Finally, we generate \textbf{entity embeddings} $\epsilon_E \in \mathbb{E}$ by applying the embedding function to all descriptions available. In case of unseen entities, the embedding can be generated on-the-fly:
\begin{equation}
\epsilon_E = \begin{cases}
  \frac{1}{|\mathcal{D}_E|} \sum_{s \in \mathcal{D}_E} \epsilon(s) & \text{ if } E \in \mathcal{K}\\
  \epsilon(E.\text{mention}) & \text{ if } E \notin \mathcal{K}
\end{cases}
\end{equation}

\subsubsection{Training Data}
We train the entity representation for diseases, syndromes and health problems using textual descriptions from various sources: 
Wikidata\footnote{\url{https://www.wikidata.org/wiki/Q12136}}, 
UMLS\footnote{\url{https://uts.nlm.nih.gov}}, 
GARD\footnote{\url{https://rarediseases.info.nih.gov}}, 
Wikipedia abstracts, 
and the Diseases Database\footnote{\url{http://www.diseasesdatabase.com}}.
In total, the knowledge base contains over 27,000 entities identified by their Wikidata ID. We trained roughly 9,700 common entities with text from Wikipedia abstracts, while we used for rare entities only their name and short description texts.

\subsection{Aspect Space}
\label{sec:discourse_aspect}

The second part of our problem is to represent the aspect in the query tuple. Here, we expect a wide range of clinical facets and we do not want to limit the users of our system to a specific terminology. Instead, we train a low-dimensional aspect vector space $\mathbb{A} \subset \mathbb{R}^d$ using the Fasttext skip-gram model \cite{bojanowski2017enriching} on medical Wikipedia articles. This approach places words with similar semantics nearby in vector space and allows queries with morphologic variations using subword representations. 

To find all possible aspects, we adopt prior work \cite{arnold2019sector} and collect all section headings from the medical Wikipedia articles. These headings typically consist of 1--3 words and describe the main topic of a section. We apply moderate preprocessing (lowercase, remove punctuation, split at ``and|\&'') to generate \textbf{aspect embeddings} $\alpha_A \in \mathbb{A}$ using a BLSTM encoder $\alpha(s)$ with the same architecture as discussed above:
\begin{equation}
\alpha_A = \frac{1}{|\mathcal{D}_A|} \sum_{s \in \mathcal{D}_A} \alpha(s)
\end{equation}

We train the embedding with over 577K sentences from Wikipedia (see Table \ref{tab:headings}). We observe that there is a vocabulary mismatch in the headings so that potentially synonymous aspects are frequently labeled with different headings, e.g.  \textsl{``types''} / \textsl{``classification''} or \textsl{``signs''} / \textsl{``symptoms''} / \textsl{``presentation''} / \textsl{``characteristics''}. However, it is also possible that in some contexts these aspects are hierarchically structured, e.g. \textsl{``presentation''} refers to the visible forms of a symptom. Our vector-space representation reflects these similarities, so it is possible to distinguish between these nuances at query time. 

\begin{table}[t]
\caption{Distribution of the top 32 headings (60\% of 577K total occurrences) contained in our training set. Numbers are given as cumulative sum. We observe that these headings cover the most important aspects for differential diagnosis.}
\begin{tabular}{@{}lrlr@{}}
\cmidrule[1pt](r{0.125em}){1-2}%
\cmidrule[1pt](lr{0.125em}){3-4}%
\textbf{Heading} & \multicolumn{1}{l}{\textbf{cusum}} & \multicolumn{1}{l}{\textbf{Heading}} & \textbf{cusum} \\

\cmidrule[0.4pt](r{0.125em}){1-2}%
\cmidrule[0.4pt](lr{0.125em}){3-4}%

information (abstract) & 9.6\% & mechanism & 54.1\% \\
treatment & 16.0\% & culture & 54.7\% \\
diagnosis & 22.3\% & society & 55.3\% \\
symptoms & 28.2\% & research & 55.9\% \\
signs & 33.0\% & risk factors & 56.4\% \\
causes & 37.1\% & presentation & 56.9\% \\
history & 39.3\% & differential diagnosis & 57.3\% \\
pathophysiology & 41.4\% & surgery & 57.7\% \\
management & 43.4\% & treatments & 58.1\% \\
epidemiology & 45.4\% & pathogenesis & 58.4\% \\
cause & 47.3\% & medications & 58.7\% \\
classification & 48.9\% & complications & 59.0\% \\
prognosis & 50.4\% & characteristics & 59.3\% \\
prevention & 51.7\% & medication & 59.5\% \\
types & 52.5\% & other animals & 59.7\% \\
genetics & 53.3\% & pathology & 60.0\% \\ 

\cmidrule[1pt](r{0.125em}){1-2}%
\cmidrule[1pt](lr{0.125em}){3-4}%

\end{tabular}
\label{tab:headings}
\end{table}

\subsection{Query Representation}
\label{sec:query_representation}
Finally, we represent the query as a tuple of entity and aspect embeddings using vector concatenation ($\oplus$):
\begin{equation}
\label{eq:query}
\begin{alignedat}{2}
Q(E,A) &= \langle\epsilon_Q \in \mathbb{E},\alpha_Q  \in \mathbb{A}\rangle = \epsilon_Q \oplus \alpha_Q 
\end{alignedat}
\end{equation}

This query encoder constitutes the upper part of our dual encoder architecture shown in Figure \ref{fig:architecture}. In the next section, we introduce our document representation that completes the lower part.

\section{Contextualized Discourse Vectors}
\label{sec:model}

\begin{figure}[t]
\includegraphics[width=0.9\columnwidth]{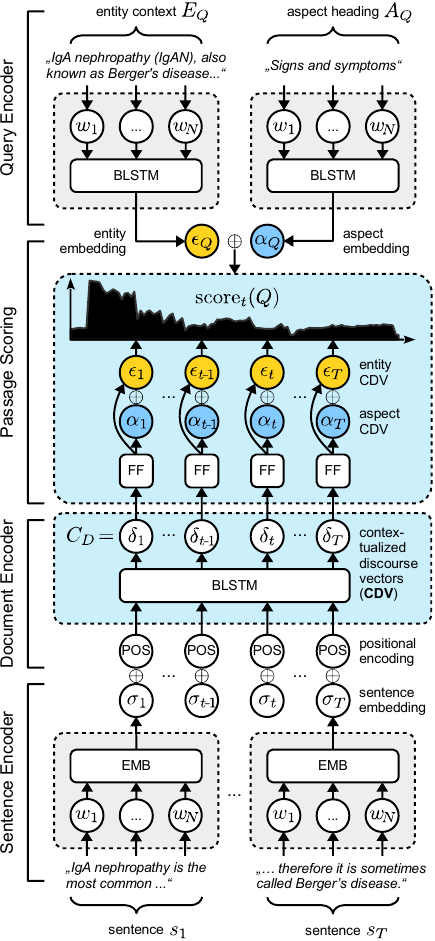}
\Description[Architecture diagram showing how documents and queries are encoded and scored.]{Architecture diagram flowing from bottom to top. On the bottom, each word (w) of a sentence in the document is encoded into a sentence embedding (sigma). These vectors are fed into a BLSTM layer that produces contextualized discourse vectors (CDV, delta). The CDV vectors are split by a feed-forward network into entity (epsilon) and aspect (alpha) embeddings. On the top, entity context is encoded into an entity query embedding (epsilon Q) and an aspect heading is encoded into an aspect query embedding (alpha Q). In the center, passages are scored by comparing both sides for each sentence in a document from left to right.}
\caption{Neural network architecture for our contextualized document representation. The contextual discourse vectors (CDV) are generated by a hierarchical stack of layers: sentence encoder (GloVe/Fasttext/BioBERT) and document encoder (BLSTM). The query encoder (entity/aspect embeddings) is used for scoring on sentence level.}
\label{fig:architecture}
\end{figure}

In this section, we introduce Contextual Discourse Vectors (CDV), a distributed document representation that focuses on coherent encoding of local discourse in the context of the entire document. The architecture of our model is shown in Figure \ref{fig:architecture}. We approach the challenges introduced in Section \ref{sec:introduction} by reading a document at word and sentence level (Section \ref{sec:model_sentence}) and encoding sentence-wise representations using recurrent layers at document level (Section  \ref{sec:model_document}). We use the representations to measure similarity between every position in the document and the query (Section \ref{sec:model_scoring}). Our model is trained to match the entity/aspect vector spaces introduced in Section \ref{sec:discourse} using self-supervision (Section \ref{sec:model_training}).

\subsection{Sentence Encoder}
\label{sec:model_sentence}

The first group of layers in our architecture encodes the plain text of an entire document into a sequence of vector representations. 
As we expect long documents---the average document length in our test sets is over 1,200 words---we chose to reduce the computational complexity by encoding the document discourse at sentence-level. It is important to avoid losing document context and word--discourse interactions (e.g. entity names or certain aspect-specific terms) during this step. Furthermore, our challenge of \emph{domain adaptability} requires the sentence encoder to be robust to linguistic and semantic shifts from text sources that differ from the training data.

Therefore we start at the input layer by encoding all words in a document $D$ into fixed low-dimensional word vectors $w_{1\dots N} \in \mathbb{R}^d$ using pre-trained word embeddings with subword information (see below). Next, we encode all sentences $s_{1\dots T} \in D$ into sentence representations $\sigma_t \in \mathbb{R}^d$ based on the words $w_k \in s_t$ in the sentence. This will reduce the number of computational time steps from $N$ words in a document to $T$ sentences. We compare two approaches for this sentence encoding step:

\subsubsection{Compositional Sentence Embeddings}

As the simplest approach we use an average vector composition of the word embeddings $w_k$ from GloVe \cite{pennington2014glove} or Fasttext \cite{bojanowski2017enriching}, which is more robust against out-of-vocabulary errors:
\begin{equation}
\sigma_{\text{avg}}(s) = \frac{1}{\text{len}(s)} \sum_{w_k \in s} w_k
\end{equation}

\subsubsection{Pooling-based Sentence Embeddings}

Since we want the model to be able to focus on individual words, we apply a language model encoder. We use the recent BioBERT \cite{lee2019biobert}, a transformer model which is pre-trained with a large amount of biomedical context on sub-word level. To generate sentence vectors from the input sequence, we use pooling of the attention layers per sentence:
\begin{equation}
\sigma_{\text{pool}}(s) = \text{BioBERT}(w_{s\text{.begin}},\dots,w_{s\text{.end}})
\end{equation}

Finally, we concatenate a positional encoding to the sentence embeddings, which encodes some rule-based structural flags such as begin/end-of-document, begin/end-of-paragraph, is-list-item. This encoding helps to guide the document encoder through the structure of a document.

\begin{figure*}[t]
\includegraphics[width=0.98\textwidth]{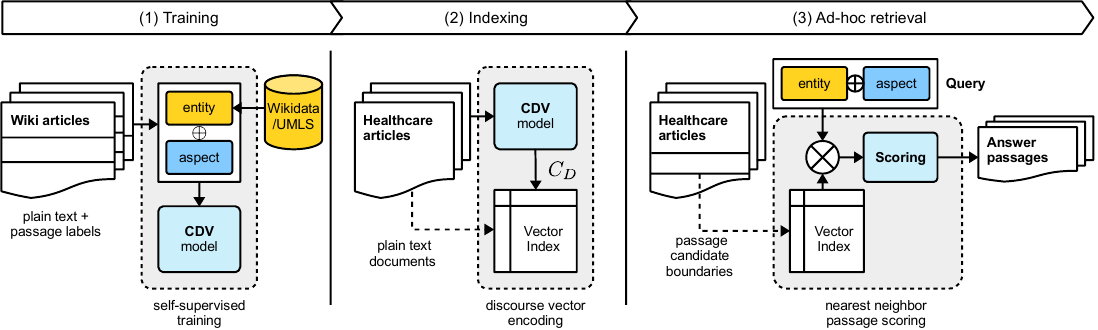}
\Description{Data flow diagram that visualizes the three stages training, indexing and ad-hoc retrieval.}
\caption{The entire answer passage retrieval process with three stages. (1) We train the discourse-aware document representation model using self-supervision on Wikipedia articles. (2) At indexing time, the model is applied once to the entire test corpus of unseen healthcare articles. The discourse vectors are saved into a vector index. (3) A query is retrieved by ranking similarity scores between the query representation and all sentence-level discourse vectors in the candidate passages.}
\label{fig:retrieval}
\end{figure*}

\subsection{Document Encoder}
\label{sec:model_document}

The second group of layers in our architecture encodes the sequence of sentences over the document. The objective of these layers is to transform the word/sentence input space into discourse vector space---which will later match with query entity and aspect spaces---in the context of the document. To achieve \emph{contextual coherence}, we use the entire document as input for a recurrent neural network with parameters $\Theta$, which we optimize at training time to minimize the loss over the sequence:
\begin{equation}
\mathcal{L}_\text{doc}(\Theta) = -\text{log} \prod_{t=1}^{T} p_\Theta\bigl(\epsilon(s_t), \alpha(s_t) \mid \sigma(s_1),\dots,\sigma(s_T)\bigr)
\end{equation}

We adopt the architecture of SECTOR \cite{arnold2019sector} and use bidirectional LSTMs to read the document sentence-by-sentence. We use a final dense layer (matrix $W_{he}$ and bias $b_e$) to produce the \textbf{local discourse vectors} $\delta_{1\dots T}$ for every sentence in $D$.
\begin{equation}
\begin{alignedat}{2}
\vec{h}_t &= \text{LSTM}_\Theta\bigl(\vec{h}_{t-1}, \sigma(s_t)\bigr) \\
\cev{h}_t &= \text{LSTM}_\Theta\bigl(\cev{h}_{t+1}, \sigma(s_t)\bigr) \\
\delta_t &= \text{tanh}\bigl(W_{he}(\vec{h}_t \oplus \cev{h}_t) + b_e\bigr)
\end{alignedat}
\end{equation}

The CDV matrix $C_D = [ \delta_1, \dots, \delta_T ]$ is our discourse-aware document representation which embeds all information necessary to decode contextualized entity and aspect information for $D$.

\subsection{Passage Scoring}
\label{sec:model_scoring}

The center layer in our architecture addresses our main task objective: find the passages with highest similarity to the query. In Section \ref{sec:discourse}, we described generalized vector spaces for entities $\epsilon \in \mathbb{E}$ and aspects $\alpha \in \mathbb{A}$ that we use as query representation $Q$ for high \emph{task coverage}. We train our discourse vectors $C_D$ to share the same vector spaces $\mathbb{E}$ and $\mathbb{A}$. This enables us to run \emph{efficient neural information retrieval} of multiple ad-hoc queries $Q$ over the pre-computed CDV vectors $C_D$ later without having to re-run inference on the document encoder for each query. We store all vectors $\delta_t$ in an in-memory vector index that allows us to efficiently retrieve approximate nearest neighbors using cosine distance. Figure \ref{fig:retrieval} shows the overall process of training, indexing and ad-hoc answer retrieval. Because we reuse entity and aspect embeddings for training, our document model `inherits' the properties from these spaces, e.g. robustness for unseen and rare entities or aspects.

\subsubsection{Discourse Decoder}
To decode the individual entity and aspect predictions $\hat\epsilon_t, \hat\alpha_t$ from $\delta_t \in C_D$, we utilize two learned decoder matrices $W_{\delta \epsilon}, W_{\delta \alpha}$ with bias terms $b_\epsilon, b_\alpha$. We optimize these parameters by using a multi-task objective with shared weights \cite{caruana1997multitask} to minimize the distance to the training labels $\epsilon_t, \alpha_t$:
\begin{equation}
\begin{alignedat}{2}
\hat\epsilon_t &= \text{tanh}(W_{\delta \epsilon} \delta_t + b_\epsilon) \\
\hat\alpha_t &= \text{tanh}(W_{\delta \alpha} \delta_t + b_\alpha) \\
\mathcal{L}_\text{cdv}(\Theta) &= \frac{1}{T} \sum_{t=1}^{T} \bigl( \lVert\hat\epsilon_t - \epsilon_t\rVert + \lVert\hat\alpha_t - \alpha_t\rVert \bigr)
\end{alignedat}
\end{equation}

\subsubsection{Sentence Scoring}

To compute similarity scores at query time, we pick up our query representation (Eq. \ref{eq:query}) and compute the semantic similarity between $Q$ and each contextual discourse vector $\delta_t$ in the vector index. To achieve low latency, we use cosine similarity between the decoded entity and aspect representations:
\begin{equation}
\begin{alignedat}{2}
\text{score}_t\bigl(Q(E,A), C_D\bigr) &= \text{cosine}\bigl(\epsilon_Q \oplus \alpha_Q, \hat\epsilon_t \oplus \hat\alpha_t\bigr) \\
&= \frac{(\epsilon_Q \oplus \alpha_Q)^\text{T} (\hat\epsilon_t \oplus \hat\alpha_t)}
{\lVert\epsilon_Q \oplus \alpha_Q\rVert \lVert\hat\epsilon_t \oplus \hat\alpha_t\rVert}
\end{alignedat}
\end{equation}

\subsubsection{Answer Retrieval}

\begin{figure}[t]
\includegraphics[width=1.0\columnwidth]{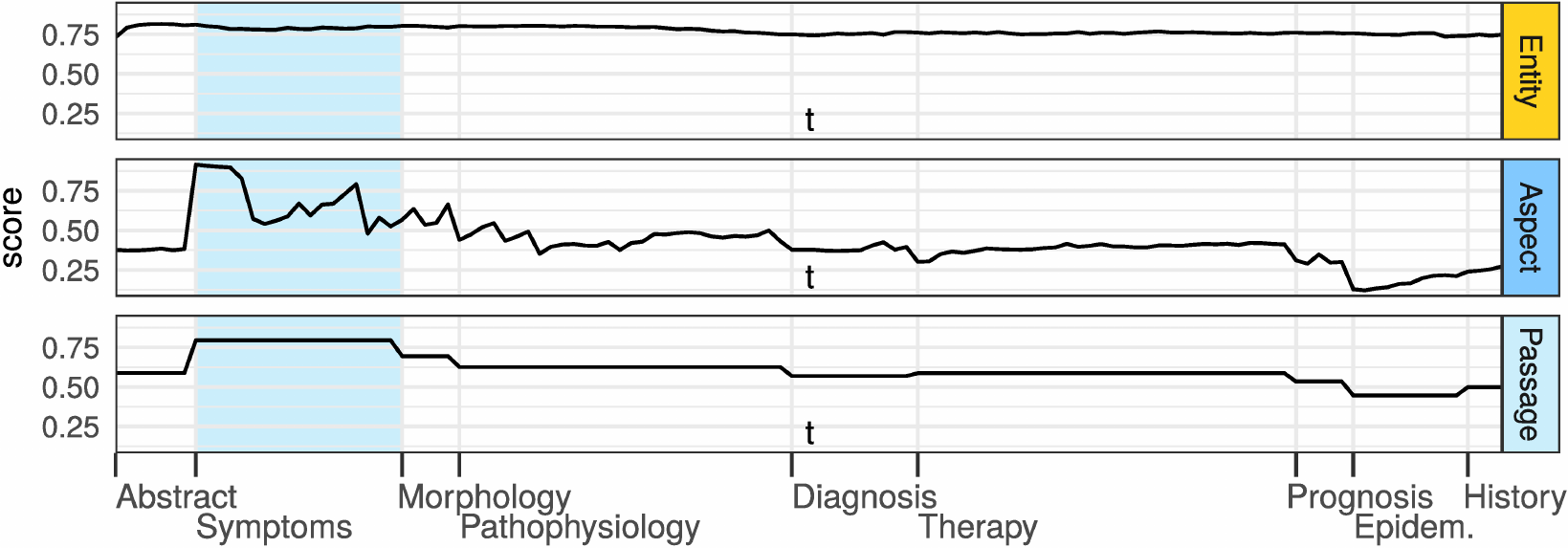}
\Description[Three plots that show the similarity of entity, aspect and passage with the query.]{Three plots that show the similarity of entity, aspect and passage (combined) of each sentence with the query. The area with highest similarity is highlighted.}
\caption{CDV model predictions for query ``symptoms of IgA nephropathy'' on the example document. The histogram shows the similarity score of the discourse vector with the query over sentences $t=1\dots T$ from left to right.}
\label{fig:histogram}
\end{figure}

The scoring operation yields a sentence-level histogram $\text{score}_{1\dots T}(Q,C_D) \in [0, 1]$ which describes the similarity between query and every sentence in a document. At this point, we have the opportunity to select a coherent set of sentences as answers similar to \cite{arnold2019sector}. However, because all healthcare datasets that we use for evaluation in this paper already provide passage boundaries, we leave this step for future work. Instead, we use the average sentence score per passage for answer retrieval:
\begin{equation}
\begin{alignedat}{2}
\text{score}(Q, P) &= \frac{1}{\text{len}(P)} \sum_{s_t \in P} \text{score}_t(Q, C_D)
\end{alignedat}
\end{equation}
Figure \ref{fig:histogram} shows the scoring curves divided into entity $Q(E)$, aspect $Q(A)$ and average $\text{score}\bigl(Q(E,A),P\bigr)$. It is clearly visible that the model coherently predicts long-range dependencies for the entity \textsl{``IgA nephropathy''} over the entire document. The aspect similarity with \textsl{``symptoms''} is much more focused on single sentences.

\subsection{Self-supervised Training}
\label{sec:model_training}

We train a generalized CDV model for all evaluation tasks by jointly optimizing all model parameters from sentence encoder, document encoder and passage scoring layers on a training set. For this task, we use the textual data about diseases and health problems available from Wikipedia. This process is self-supervised, because there exist no labeled query-answer pairs for these documents. Instead, we assign for each sentence $s_t \in D$ a set of related entities $E$ and aspects $A$ using simple heuristics:
\begin{equation}
\begin{alignedat}{2}
E(s_t, D) &= \{E \mid \text{title}(D) = E \vee \text{contains\_link}(s_t, E)\}\\
A(s_t, D) &= \{A \mid \text{heading}(s_t) = A\}
\end{alignedat}
\end{equation}

We collected over 8,600 articles for training and removed all instances contained in any of the test sets. The collection contains information about over 8K entities and 15K aspects (see Table \ref{tab:datasets}).

\subsubsection{Discourse Encoding}
We create the target objectives for training using the average of the label embeddings contained in the training entities $E(s_t, D)$ and aspects $A(s_t, D)$ on sentence level:
\begin{eqnarray}
\begin{alignedat}{2}
\epsilon_t &= \frac{\sum_{E \in E(s_t, D)} \epsilon_E}{| E(s_t, D) |}\\
\alpha_t &= \frac{\sum_{A \in A(s_t, D)} \alpha_A}{| A(s_t, D) |}
\end{alignedat}
\end{eqnarray}

\subsubsection{Optimized Loss Function}

We observe a strong imbalance of entity and aspect labels over the course of a single document, for example when passages contain lists (very short sentences), rare entities or have uncommon headlines. To give the network the ability to capture these anomalies, especially with larger batch sizes, we use a robust loss function \cite{barron2019general} which resembles a smoothed form of Huber Loss \cite{huber1992robust}:
\begin{eqnarray}
\mathcal{L}_\text{cdv+}(\Theta) = \frac{1}{T} \sum_{t=1}^{T} \left( \sqrt{1 + \left(\frac{\lVert\hat\epsilon_t - \epsilon_t\rVert + \lVert\hat\alpha_t - \alpha_t\rVert}{4}\right)^2} - 1 \right)
\end{eqnarray}

In the next section, we apply our CDV model to a healthcare answer retrieval task.

\section{Experimental Results}
\label{sec:evaluation}

We evaluate our CDV model and 14 baseline methods in an answer passage retrieval task. All models are trained using self-supervision on Wikipedia texts and applied as zero-shot task \cite{palatucci2009zero} (i.e. without further fine-tuning) to three diverse English healthcare datasets \emph{WikiSection}, \emph{MedQuAD} and \emph{HealthQA}. 

\subsection{Evaluation Setup}

As queries, we use tuples of the form $\langle\text{entity}, \text{aspect}\rangle$. Because our task requires to retrieve the answers from over 4,000 passages and the interaction-based models in our comparison require computationally expensive pairwise inference, we evaluate all numbers on a re-ranking task \cite{gillick2018end}. We follow the setup of \citet{logeswaran2019zero} and use BM25 \cite{robertson1995okapi} to provide each model with a pre-filtered set of 64 potentially relevant passage candidates\footnote{This choice covers 80-91\% of all true answers (depending on the dataset) as trade-off between task complexity and real-world applicability. The numbers reported for HealthQA in the original paper were evaluated by re-ranking ten candidates (one relevant, 3 partially relevant and 6 irrelevant) and are therefore not comparable.}. To facilitate full recall in this model comparison, we add missing true answers to the candidates if necessary by overwriting the lowest-ranked false answers in the list and shuffle afterwards. We rank the candidate answers using exhaustive nearest neighbor search and leave the evaluation of indexing efficiency for future work. Next, we describe the datasets, metrics and methods used in our experiments.

\begin{table}[t]
\caption{Statistics of our training and evaluation data sets.}
\begin{tabular}{@{}lrlrrr@{}}
\toprule
\textbf{Dataset} & \multicolumn{1}{c}{\textbf{Wikipedia}} & $\Rightarrow$ & \textbf{WS} & \textbf{MQ} & \textbf{HQ} \\  
Split & \textbf{train} &  & \textbf{test} & \textbf{test} & \textbf{test} \\
\cmidrule(r){1-2} \cmidrule(l{0.5em}){4-6}
\# documents & 8,605 & \textbf{} & 716 & 1,111 & 178 \\
\# passages & 53,477 & \textbf{} & 4,373 & 3,762 & 1,109 \\
\# entities & 8,605 & \textbf{} & 716 & 1,100 & 221 \\
\# aspects & 15,028 & \textbf{} & 27 & 15 & 21 \\
\# queries & \textbf{N/A} & \textbf{} & 4,178 & 3,294 & 1,045 \\ 
\cmidrule(r){1-2} \cmidrule(l{0.5em}){4-6}
avg words/doc & 977.6 &  & 1,396.7 & 811.1 & 1,449.4 \\
avg sents/doc & 43.5 &  & 63.7 & 48.0 & 82.5 \\
avg passages/doc & 6.2 &  & 6.1 & 3.4 & 6.2 \\
avg words/passage & 221.8 &  & 228.6 & 237.9 & 232.6 \\
avg sents/passage & 9.8 &  & 10.4 & 14.1 & 13.2 \\
avg words/sent & 22.7 &  & 21.9 & 16.9 & 17.6 \\ \bottomrule
\end{tabular}
\label{tab:datasets}
\end{table}

\begin{table*}[t]
\caption{Results for the answer passage retrieval task on three healthcare datasets. All models were trained using the self-supervised Wikipedia training set and applied without fine-tuning. Queries were evaluated by ranking 64 given candidates from the respective test sets. As queries we used $\langle\text{entity}, \text{aspect}\rangle$ tuples in a representation suitable for the individual model.}
\begin{tabular}{@{}lrrrrrrrrr@{}}
\toprule
\textbf{Model} & \multicolumn{3}{c}{\textbf{WikiSectionQA}} & \multicolumn{3}{c}{\textbf{MedQuAD}} & \multicolumn{3}{c}{\textbf{HealthQA}} \\
all trained on Wikipedia & $R@1$ & $R@10$ & MAP & $R@1$ & $R@10$ & MAP & $R@1$ & $R@10$ & MAP \\

\cmidrule[0.4pt](r{0.125em}){1-1}%
\cmidrule[0.4pt](lr{0.125em}){2-4}%
\cmidrule[0.4pt](lr{0.125em}){5-7}%
\cmidrule[0.4pt](lr{0.125em}){8-10}%
\emph{Term-based models} &&&&&&&&& \\
TF-IDF \cite{sparck1972statistical} & 17.10 & 64.99 & 31.77 & 23.83 & 82.84 & 42.66 & 17.46 & 71.54 & 34.47 \\
BM25 \cite{robertson1995okapi} & 23.87 & 71.26 & 38.89 & 29.48 & 86.11 & 48.89 & 22.55 & 73.27 & 38.45 \\
\emph{Representation-based models} &&&&&&&&& \\
ARC-I \cite{hu2014convolutional} & 1.61 & 13.69 & 6.90 & 1.98 & 19.22 & 8.47 & 1.38 & 13.87 & 6.87 \\
DSSM \cite{huang2013learning} & 22.82 & 74.31 & 39.02 & 13.11 & 55.92 & 27.38 & 10.50 & 46.44 & 22.04 \\
C-DSSM \cite{shen2014learning} & 9.59 & 53.12 & 22.82 & 9.67 & 47.54 & 22.12 & 10.56 & 58.30 & 25.37 \\
\emph{Interaction-based models} &&&&&&&&& \\
ARC-II \cite{hu2014convolutional} & 10.38 & 53.62 & 23.61 & 9.19 & 47.58 & 21.66 & 11.26 & 58.85 & 26.09 \\
DRMM \cite{guo2016deep} & 24.96 & 67.56 & 39.24 & 34.52 & 82.35 & 51.51 & 21.80 & 80.24 & 40.03 \\
MatchPyramid \cite{pang2016text} & 18.53 & 64.21 & 33.12 & 25.14 & 72.33 & 41.37 & 19.24 & 73.79 & 37.22 \\
aNMM \cite{yang2016anmm} & 4.77 & 32.17 & 14.03 & 7.15 & 37.18 & 17.08 & 3.74 & 27.20 & 12.07 \\
KNRM \cite{xiong2017end} & 16.96 & 61.03 & 31.04 & 16.86 & 61.35 & 31.35 & 22.94 & 67.92 & 37.65 \\
CONV-KNRM \cite{dai2018convolutional} & 34.36 & 77.25 & 48.72 & 42.70 & 84.54 & 57.57 & 33.13 & 85.41 & 50.55 \\
HAR \cite{zhu2019hierarchical} & 45.31 & 84.15 & 58.38 & \textbf{55.65} & \textbf{93.17} & \textbf{69.10} & 43.20 & 88.34 & 58.80 \\ 
\emph{Combined models} &&&&&&&&& \\
Duet \cite{mitra2017learning} & 18.34 & 59.13 & 31.74 & 20.50 & 65.91 & 35.28 & 17.27 & 64.81 & 32.13 \\
MVLSTM \cite{wan2016deep} & 30.74 & 76.10 & 45.58 & 36.86 & 86.29 & 53.18 & 26.78 & 84.42 & 45.37 \\

\emph{Our model} &&&&&&&&& \\
CDV+avg-glove & 59.60 & 95.67 & 72.72 & 34.00 & 80.87 & 50.45 & 37.17 & 84.47 & 53.30 \\
CDV+avg-fasttext & 60.34 & 97.49 & 74.01 & 45.26 & 92.29 & 62.56 & 40.08 & \textbf{89.80} & 58.35 \\
CDV+pool-biobert & \textbf{65.21} & \textbf{97.84} & \textbf{77.60} & 39.96 & 91.32 & 58.91 & \textbf{43.60} & 88.12 & \textbf{59.40} \\
\bottomrule
\end{tabular}
\label{tab:retrieval}
\end{table*}

\subsubsection{Evaluation Datasets}

We conduct experiments on three English datasets from the clinical and healthcare domain (see Table \ref{tab:datasets}). From the documents provided, we use the plain text of the entire document body during model inference and the segmentation information for generating the passage candidates. From all queries provided, we use the entity labels (mention text, Wikidata ID) and aspect labels (UMLS canonical name). If entity and aspect identifiers were not provided by the dataset, we added them manually by asking three annotators from clinical healthcare to label them.

\emph{WikiSectionQA} \cite{arnold2019sector} (WS) is a large subset of full-text Wikipedia articles about diseases, labeled with entity identifiers, section headlines and 27 normalized aspect classes. We extended this dataset for answer retrieval by constructing query tuples from every section containing the given entity ID and normalized aspect label. We included abstracts as \textsl{``information''}, but skipped sections labeled as \textsl{``other''}. We use the en\_disease-test split for evaluation and made sure that none of the documents are contained in our training data.

\emph{MedQuAD} \cite{abacha2019question} (MQ) is a collection of medical question-answer pairs from multiple trusted sources of the National Institutes of Health (NIH): National Cancer Institute (NCI)\footnote{\url{https://www.cancer.gov}}, Genetic and Rare Diseases (GARD) \footnote{\url{https://rarediseases.info.nih.gov}}, Genetics Home Reference (GHR)\footnote{\url{https://ghr.nlm.nih.gov}}, National Institute of Diabetes and Digestive and Kidney Diseases (NIDDK) \footnote{\url{http://www.niddk.nih.gov/health-information/health-topics/}}, National Institute of Neurological Disorders and Stroke (NINDS) \footnote{\url{http://www.ninds.nih.gov/disorders/}}, NIH Senior Health \footnote{\url{http://nihseniorhealth.gov/}} and National Heart, Lung and Blood Institute (NHLBI)\footnote{\url{http://www.nhlbi.nih.gov/health/}}. We left out documents from Medline Plus due to property rights.
Questions are annotated with structured identifiers for entities (UMLS CUI), aspect (semantic question type) and contain a long passage as answer. To make this dataset applicable to our method, we reconstructed the entire documents from the answer passages and kept only questions about diseases for evaluation. We filtered out documents with only one passage (these were always labeled ``information'') and separated a random 25\% test split from the remaining documents.

\emph{HealthQA} \cite{zhu2019hierarchical} (HQ) is a collection of consumer health question-answer pairs crawled from the website Patient\footnote{\url{https://patient.info}}. The answer passages were generated from sections in the documents and annotated by human labelers with natural language questions. We reconstructed the full documents from these sections. Additionally, our annotators added structured entity and aspect labels to all questions in the test split. Although some questions are not about diseases, we kept all of them to remain comparable with related work.

\subsubsection{Evaluation Metrics}

For all ranking experiments, we use \emph{Recall at top K} (R@K) and \emph{Mean Average Precision} (MAP) metrics. While R@1 measures if the top-1 answer is correct or not (similar to a question answering task), we also report R@10, which corresponds with the ability to retrieve all correct answers in a top-10 results list, and MAP, which considers the entire result list.

\subsubsection{Baseline Methods}

We evaluate two term-based matching functions as baseline: TF-IDF \cite{sparck1972statistical} and BM25 \cite{robertson1995okapi}. We used the implementation in Apache Lucene 8.2.0\footnote{\url{https://lucene.apache.org}} to retrieve passages containing entity and aspect of a query, e.g. \textsl{``IgA nephropathy symptoms''} from the index of all passages in the test dataset.

Additionally, we evaluate the following document matching methods from the literature: 
ARC-I and ARC-II \cite{hu2014convolutional}, 
DSSM \cite{huang2013learning}, 
C-DSSM \cite{shen2014learning}, 
DRMM \cite{guo2016deep},
MatchPyramid \cite{pang2016text},
aNMM \cite{yang2016anmm},
Duet \cite{mitra2017learning},
MVLSTM \cite{wan2016deep},
KNRM \cite{xiong2017end},
CONV-KNRM \cite{dai2018convolutional} and
HAR \cite{zhu2019hierarchical}. 
For implementing these models, we followed \citet{zhu2019hierarchical} and used the open source implementation MatchZoo \cite{guo2019matchzoo} with pre-trained glove.840B.300d vectors \cite{pennington2014glove}. All models were trained with our self-supervised Wikipedia training set using queries containing the entity and lowercase heading, e.g. \textsl{``IgA nephropathy ; symptoms''} and applied to the test sets using queries of the same structure, instead of natural language questions.

\subsection{Implementation Details}

We implement our models with the following configurations:
For the sentence encoding, we use either glove.6B.300d pre-trained GloVe vectors (+avg-glove), 128d fine-tuned Fasttext embeddings (+avg-fasttext) or the 768d pre-trained BioBERT \cite{lee2019biobert} language model (+pool-biobert).
For the document encoding, we use two LSTM layers (one forward, one backward) with 512 dimensions each, a discourse vector dense layer with 256 dimensions, L2 batch normalization and tanh activation. 
The discourse decoder is a 128-dimensional output layer with tanh activation and Huber loss. 
The network is trained with stochastic gradient descent over 50 epochs using the ADAM optimizer with a batch size of 16 documents, a learning rate of $10^{-3}$, $0.975$ exponential decay per epoch, $0.0$ dropout and $10^{-4}$ weight decay regularization. We chose these parameters using hyperparameter search on the WikiSection validation set. During training, we restrict the maximum document length to 396 sentences and maximum sentence length to 96 tokens, due to memory constraints on the GPU.

The entity and aspect embeddings are trained with 128d Fasttext embeddings, followed by BLSTM and dense embedding layers, each with 128 dimensions and tanh activations. The output layer is configured with 1024 dimensions, sigmoid activation and BPMLL loss \cite{zhang2006multilabel} to predict the Bloom filter hash ($k=5$) of the entity or aspect. The network is trained similarly to the CDV model, but we use 5 epochs with a batch size of 128 sentences, a learning rate of $10^{-3}$ and $0.5$ dropout.

\subsection{Results}

\begin{figure}[t]
\includegraphics[width=1.0\columnwidth]{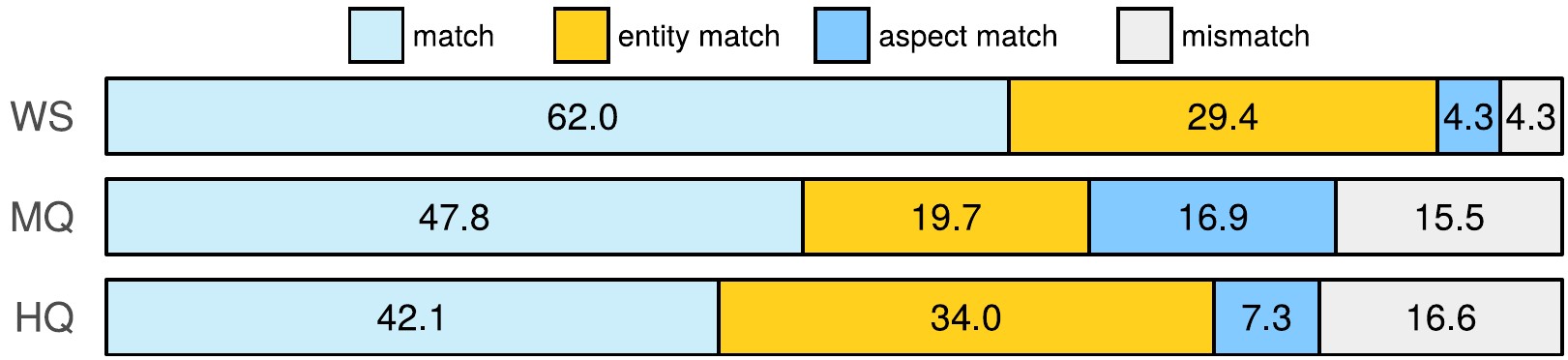}
\Description[Plot showing the proportions for match, entity match, aspect match and mismatch.]{Plot showing the proportions for match (42--62\%), entity match (20--34\%), aspect match (4--17\%) and mismatch (4--17\%) for three datasets.}
\caption{Entity/aspect matching (values in \%) observed on all examples in the three evaluation datasets.}
\label{fig:entityaspecterrors}
\end{figure}

Table \ref{tab:retrieval} shows the results on the answer passage retrieval task using CDV and document matching models on three healthcare datasets.
We observe that CDV consistently achieves significantly better results than all term-based, representation-based and combined models across all datasets. In comparison with pairwise interaction-based models, our representation-based retrieval model outperforms all tested models on average, scores best on WikiSection and HealthQA and second best on the MedQuAD dataset. Retrieval time per query is 247ms ($\pm$43ms) on average. Figure \ref{fig:entityaspecterrors} further shows that we correctly match between 67.5\% and 91.4\% of entities in the datasets and resolve 49.4\% to 66.3\% of all aspects.

\paragraph{Comparison of Model Architectures}

Our query model is able to match most of the questions in the entity/aspect scheme (see Section \ref{sec:discussion} for exceptions). 
The results show that term-based TF-IDF and BM25 models can solve the healthcare retrieval task sufficiently with R@10 $> 70\%$. In contrast, none of the representation-based re-ranking models can achieve similar performance, except DSSM on WikiSectionQA. 
Most of the recent interaction-based and combined models outperform BM25 and have significant advantages on the MedQuAD dataset, which contains a large amount of generated information that can be matched exactly. We follow that simple word-level interactions are important for this task and representation-based models trade off this property for fast retrieval times. 

\paragraph{Background Knowledge} 

Our CDV model performs well on all data sets, but shows a significant advantage on the Wikipedia-based WikiSectionQA dataset. Although all models are trained on the same data, the only model with similar behavior is DSSM. One possible reason is that entity embeddings are an important source for background information, and these are mainly based on Wikipedia descriptions. 99.9\% of the WikiSectionQA entities are covered in our embedding, 97.1\% on MedQuAD and only 69.29\% on HealthQA, because it does not only contain diseases. 
Additionally, sentence embeddings provide different levels of background knowledge and language understanding. The pre-trained GloVe embedding can handle the task well, but is outperformed by our fine-tuned Fasttext embedding and the large BioBERT language model.

\paragraph{Domain Adaptability and Task Coverage}

\begin{figure}[t]
\includegraphics[width=1.0\columnwidth]{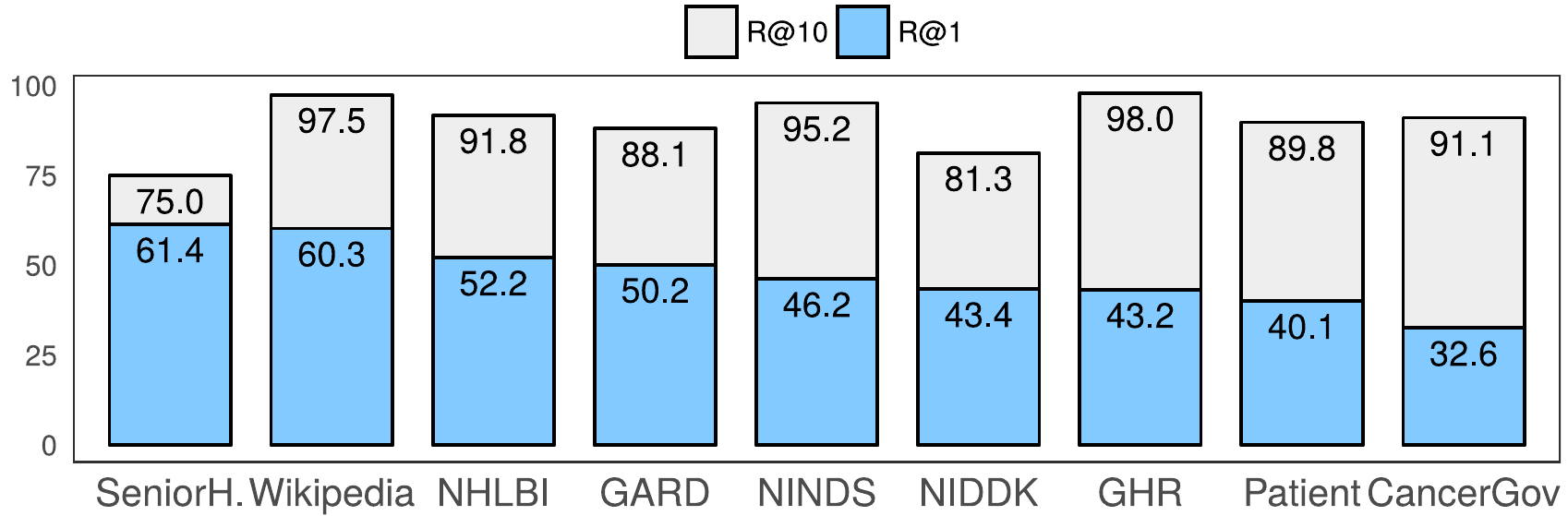}
\Description[Bar plot showing the individual Recall values for each data source.]{Bar plot showing the individual Recall@1/10 values for data sources Senior Health (61/75\%), Wikipedia (60/98\%), NHBLI (52/92\%), GARD (50/88\%), NINDS (46/95\%), NIDDK (43/81\%), GHR (43/98\%), Patient (40/90\%) and CancerGov (33/91\%).}
\caption{R@1 and R@10 performance of the CDV-EA+pool-biobert model across all data sources.}
\label{fig:datasources}
\end{figure}

\begin{figure}[t]
\includegraphics[width=1.0\columnwidth]{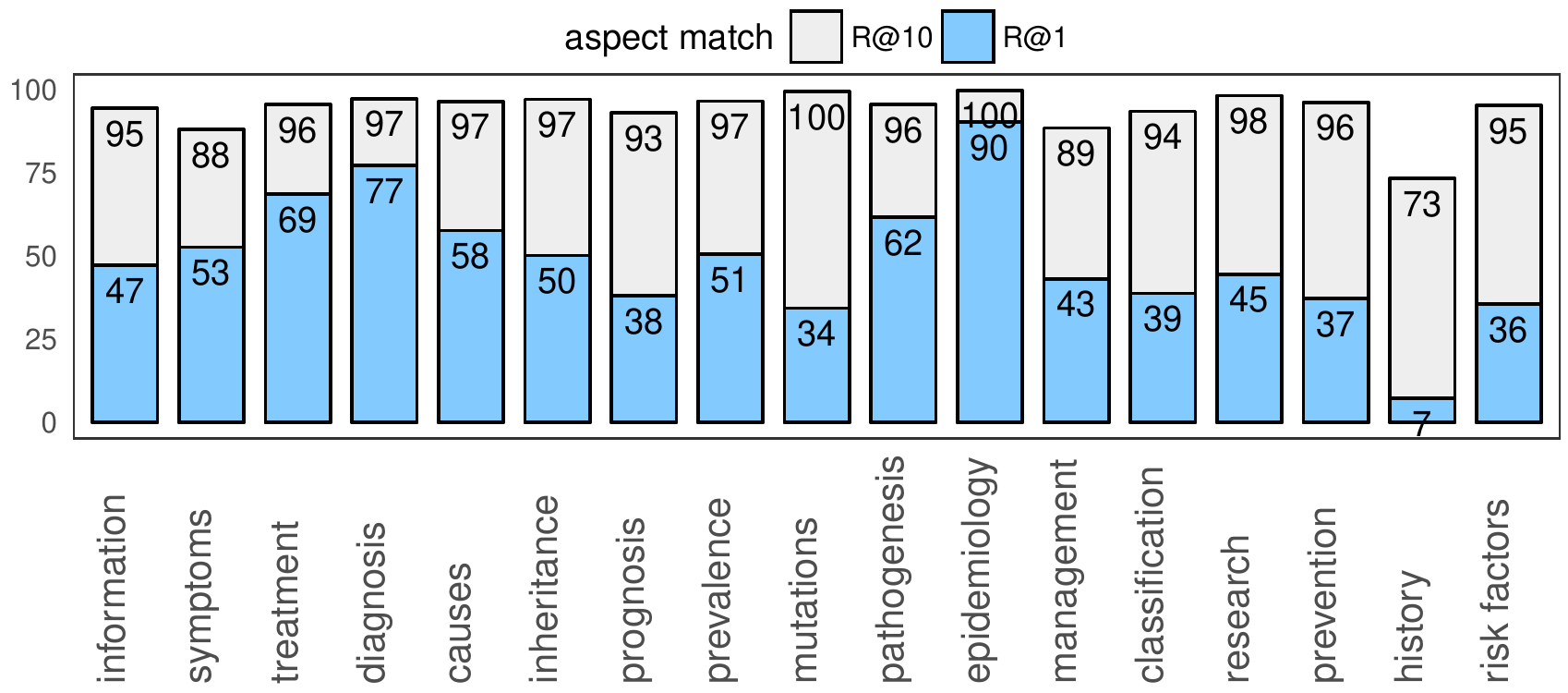}
\Description[Bar plot showing the individual Recall values for each aspect.]{Bar plot showing the individual Recall@1 values aspects information (47\%), symptoms (53\%), treatment (69\%), diagnosis (77\%), causes (58\%), inheritance (50\%), prognosis (38\%), prevalence (51\%), mutations (34\%), pathogenesis (62\%), epidemiology (90\%), management (43\%), classification (39\%), research (45\%), prevention (37\%), history (7\%) and risk factors (36\%). Recall@10 varies between 89--100\% except for history with 73\%.}
\caption{R@1 and R@10 prediction performance on the 17 most frequent aspects in all test sets (of 34 total).}
\label{fig:aspectclasses}
\end{figure}

Figure \ref{fig:datasources} shows the performance of our CDV+avg-fasttext model across all data sources, most of them contained in MedQuAD. This distribution reveals that our model top-1 accuracy is stable in the adaptation to most sources except National Cancer Institute (CancerGov). However, we notice that R@10 performance is high among all sources except SeniorHealth. Figure \ref{fig:aspectclasses} shows that R@10 performance across the most frequent aspects is over 93\% in most cases, but with varying top-1 recall. We will address these errors in Section \ref{sec:discussion}.

\paragraph{Impact of Contextual Dependencies}

An important feature of our CDV model is that all score predictions are calculated on sentence level with respect to long-range context across the entire document. 
In Figure \ref{fig:histogram}, we observe that the model is able to predict the entity (top curve) consistently over the document, although there are many coreferences in Wikipedia text. The aspect curve (center) clearly shows the beginning of the expected section \textsl{``Symptoms''} and the model is uncertain for the following sentences.
Finally, the average score (bottom curve) shows a coherent prediction.

\subsection{Discussion and Insights}
\label{sec:discussion}

\begin{figure}[t]
\includegraphics[width=1.0\columnwidth]{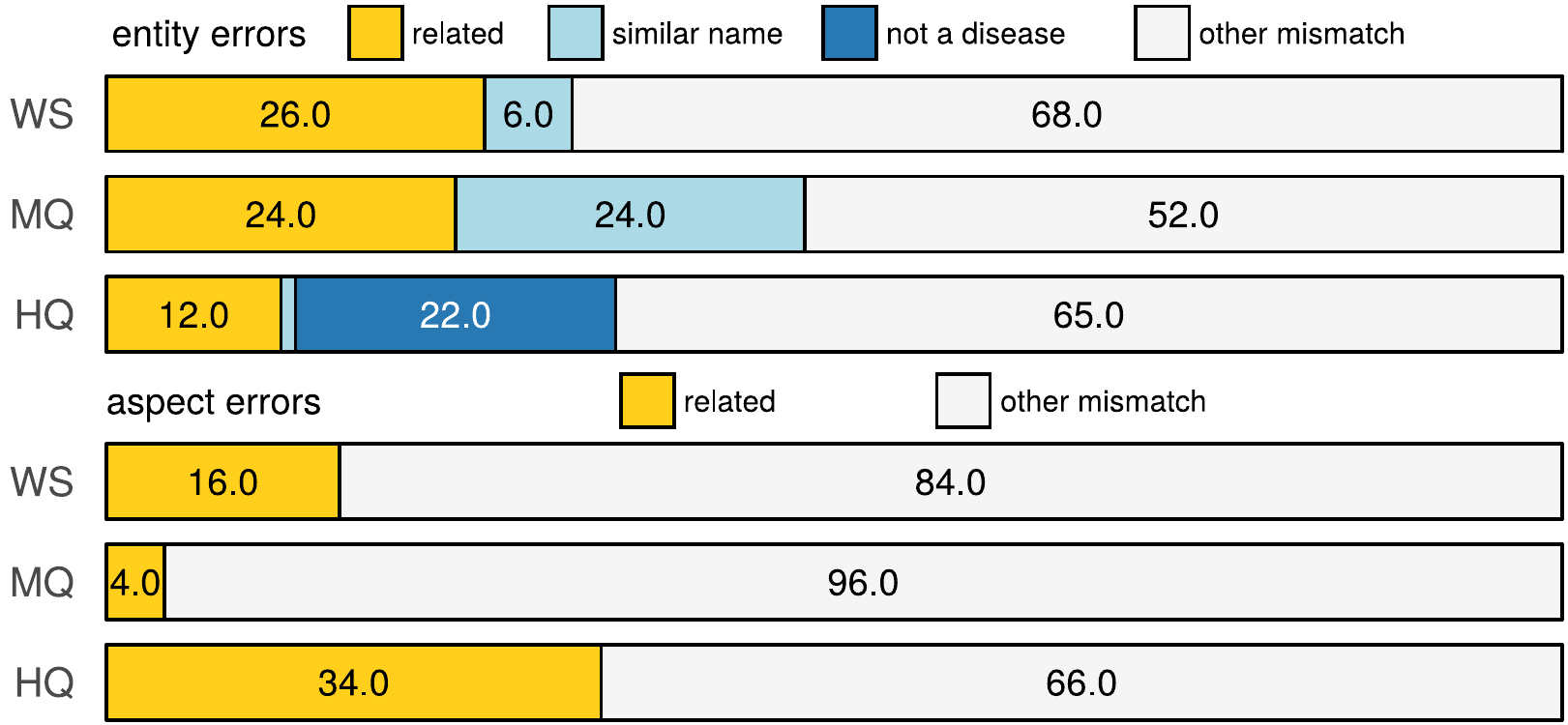}
\Description[Plot showing the proportion of entity and aspect errors among the mismatched queries.]{Plot showing the proportion of entity errors (related 12--26\%, similar name 1-24\%, not a disease 0--22\%, other mismatch) and aspect errors (related 4--34\%, other mismatch) among the mismatched queries the three datasets.}
\caption{Error classes for entity and aspect mismatch (values in \%) from manual analysis of 150 mismatched queries. }
\label{fig:errorclasses}
\end{figure}

We perform an error analysis on the predictions of the CDV+avg-fasttext model to identify main reasons for answer misranking. For this purpose we analyse samples in which the model ranks a wrong passage at the top-1 position. We look at 50 random mismatched samples per dataset to understand the individual challenges per source. We discuss the main findings in the following.

\paragraph{Related Entities}
Figure \ref{fig:errorclasses} shows that a main source of entity errors comes from selecting passages that belong to related entities. This includes entities that are superclasses or subclasses of the gold truth, e.g. selecting a passage covering \textsl{``Diarrhea''} when \textsl{``Chronic Diarrhea in Children''} is the query entity. 
These errors are most significant in WikiSectionQA and MedQuAD, because HealthQA covers mostly common diseases. We especially observe this in samples from Genetic Home Reference and National Cancer Institute. 
Figure \ref{fig:datasources} shows that R@1 is low for samples from these sources, whereas their R@10 is high. That is because genetic conditions and cancer types inherently contain entities with very similar names and descriptions. For instance, we see \textsl{``Spastic Paraplegia Type 8''} falsely resolved to \textsl{``Spastic Paraplegia Type 11''}. As the representations are close to each other in vector space, the correct samples are almost always found within the top-10 ranked candidates, corresponding with the high R@10.

\paragraph{Related Aspects}
Likewise, we observe that in HealthQA 34\% and in WikiSectionQA 16\% of aspects are mismatched to related aspects. Figure \ref{fig:aspectclasses} shows the distribution of aspects and the model's ability to resolve them. 
It is salient that some aspects are especially difficult to resolve. Aside from the fact that these aspects are in the long tail, a further analysis reveals that they are often resolved to related aspects. For example, passages covering \textsl{``classification''} are often very similar and therefore confused with passages about \textsl{``diagnosis''} and \textsl{``symptoms''}. The same holds true for \textsl{``prognosis''} and \textsl{``management''}. Queries asking for \textsl{``prevalence''} of a disease are often resolved to \textsl{``information''} passages, because disease frequency is often mentioned in these introductory texts. In general, passages about related aspects often share similar tokens and document context, which makes their distinction more difficult.

\paragraph{Out-of-Scope Questions.}
25\% of queries from HealthQA contain entities that are no diseases but procedures, drugs or other entity types. As our model is trained on textual data covering diseases only, we do not expect it to fully resolve these entities. However, we observe that the model is capable of finding the correct passage for 23\% of unseen entities. This shows that while our model is not trained on such entity types, the fallback embedding described in \ref{sec:entityembeddings} still allows to generalize even to non-diseases in these cases.

\paragraph{Evaluation vs. Real-World Application}
We further identified a number of errors related to the structure of the evaluation, that would be less problematic or even beneficial in real-world application. The model frequently ranks passages to the top which answer the query but have a differing aspect assigned. We observe this in 24\% of analysed samples from WikiSectionQA and 18\% from HealthQA. This often seems to be caused by the non-discrete nature of topical aspects. In practice a passage can cover more than one aspect, but our evaluation does currently not capture this ambiguity. Additionally we find some mismatches between passages and their ground-truth aspects, which can be ascribed to writing errors in WikiSectionQA and labeling errors in HealthQA. Aspects in MedQuAD are less ambiguous in general and only 4\% fall into this error class. Figure \ref{fig:entityaspecterrors} shows that the model therefore resolves more aspects correctly for MedQuAD queries.


\paragraph{Irrelevant Text within Passage Boundaries}
Another finding is that 28\% of analysed samples from the MedQuAD dataset contain boilerplate text unrelated to a specific entity. The boilerplate includes repeated text such as information about how data was collected. In this case our model is able to detect relevant parts of a passage (see Figure \ref{fig:histogram}), but the remaining irrelevant sentences lead to a worse ranking of the passage. Evaluating with flexible passage boundaries would eliminate this issue and be a better match for real-world scenarios, in which the interest of a medical professional is mainly focused on non-boilerplate parts of a document.

\paragraph{Complex Questions} 
We find that most questions in our evaluation can be represented as tuples of entity and aspect without information loss. However, in 4\% of analysed queries in the HealthQA dataset we see a mismatch between question and query. For instance, the question \textsl{``How common is OCD in Children and Young People?''} which is more specific than the assigned query tuple \textsl{``Obsessive-compulsive disorder''} and \textsl{``prevalence''}. Different solutions are possible for representing more complex queries, e.g. 
by composing multiple queries during retrieval. We leave these questions for future research.

\section{Conclusions and Future Work}
\label{sec:conclusion}

We present CDV, a contextualized document representation that uses structured entity/aspect queries to retrieve answer passages from long healthcare documents. Our model is based on a hierarchical dual encoder architecture which combines interaction-based feature encoding with low-latency continuous retrieval. In comparison to previous approaches, CDV is able to integrate document context into its representations, which helps to resolve long-range dependencies normally not visible to passage re-ranking models. We train a self-supervised model on medical Wikipedia texts and show that it applies to three healthcare answer retrieval tasks best or second best, compared to 14 strong baseline models. We trained all models using the same data and provide structured labels on three existing datasets for the evaluation of this task.

In future work, we will address the transfer of the CDV model to different languages, rare diseases and more complex question types. Another open challenge is the extraction of passage boundaries during retrieval. Furthermore, it is interesting to see how fine-tuning the model on supervised data will improve retrieval performance.
\begin{acks}
Our work is funded by the German Federal Ministry of Economic Affairs and Energy (BMWi) under grant agreement 01MD19003b (PLASS) and 01MD19013D (Smart-MD).
\end{acks}

\bibliographystyle{ACM-Reference-Format}
\bibliography{main}


\begin{thebibliography}{52}


\ifx \showCODEN    \undefined \def \showCODEN     #1{\unskip}     \fi
\ifx \showDOI      \undefined \def \showDOI       #1{#1}\fi
\ifx \showISBNx    \undefined \def \showISBNx     #1{\unskip}     \fi
\ifx \showISBNxiii \undefined \def \showISBNxiii  #1{\unskip}     \fi
\ifx \showISSN     \undefined \def \showISSN      #1{\unskip}     \fi
\ifx \showLCCN     \undefined \def \showLCCN      #1{\unskip}     \fi
\ifx \shownote     \undefined \def \shownote      #1{#1}          \fi
\ifx \showarticletitle \undefined \def \showarticletitle #1{#1}   \fi
\ifx \showURL      \undefined \def \showURL       {\relax}        \fi
\providecommand\bibfield[2]{#2}
\providecommand\bibinfo[2]{#2}
\providecommand\natexlab[1]{#1}
\providecommand\showeprint[2][]{arXiv:#2}

\bibitem[\protect\citeauthoryear{Abacha and Demner-Fushman}{Abacha and
  Demner-Fushman}{2019}]%
        {abacha2019question}
\bibfield{author}{\bibinfo{person}{Asma~Ben Abacha} {and} \bibinfo{person}{Dina
  Demner-Fushman}.} \bibinfo{year}{2019}\natexlab{}.
\newblock \showarticletitle{A {{Question}}-{{Entailment Approach}} to
  {{Question Answering}}}.
\newblock \bibinfo{journal}{\emph{BMC Bioinformatics}} \bibinfo{volume}{20},
  \bibinfo{number}{1} (\bibinfo{year}{2019}), \bibinfo{pages}{511}.
\newblock


\bibitem[\protect\citeauthoryear{Abacha, Shivade, and Demner-Fushman}{Abacha
  et~al\mbox{.}}{2019}]%
        {abacha2019overview}
\bibfield{author}{\bibinfo{person}{Asma~Ben Abacha}, \bibinfo{person}{Chaitanya
  Shivade}, {and} \bibinfo{person}{Dina Demner-Fushman}.}
  \bibinfo{year}{2019}\natexlab{}.
\newblock \showarticletitle{Overview of the Mediqa 2019 Shared Task on Textual
  Inference, Question Entailment and Question Answering}. In
  \bibinfo{booktitle}{\emph{Proceedings of the 18th {{BioNLP Workshop}} and
  {{Shared Task}}}}. \bibinfo{pages}{370--379}.
\newblock


\bibitem[\protect\citeauthoryear{Adolphs, Theobald, Schafer, Uszkoreit, and
  Weikum}{Adolphs et~al\mbox{.}}{2011}]%
        {adolphs2011yago}
\bibfield{author}{\bibinfo{person}{Peter Adolphs}, \bibinfo{person}{Martin
  Theobald}, \bibinfo{person}{Ulrich Schafer}, \bibinfo{person}{Hans
  Uszkoreit}, {and} \bibinfo{person}{Gerhard Weikum}.}
  \bibinfo{year}{2011}\natexlab{}.
\newblock \showarticletitle{{{YAGO}}-{{QA}}: {{Answering}} Questions by
  Structured Knowledge Queries}. In \bibinfo{booktitle}{\emph{Fifth
  {{International Conference}} on {{Semantic Computing}}}}.
  \bibinfo{publisher}{{IEEE}}, \bibinfo{pages}{158--161}.
\newblock


\bibitem[\protect\citeauthoryear{Arnold, Schneider, Cudr{\'e}-Mauroux, Gers,
  and L{\"o}ser}{Arnold et~al\mbox{.}}{2019}]%
        {arnold2019sector}
\bibfield{author}{\bibinfo{person}{Sebastian Arnold}, \bibinfo{person}{Rudolf
  Schneider}, \bibinfo{person}{Philippe Cudr{\'e}-Mauroux},
  \bibinfo{person}{Felix~A. Gers}, {and} \bibinfo{person}{Alexander
  L{\"o}ser}.} \bibinfo{year}{2019}\natexlab{}.
\newblock \showarticletitle{{{SECTOR}}: {{A Neural Model}} for {{Coherent Topic
  Segmentation}} and {{Classification}}}.
\newblock \bibinfo{journal}{\emph{Transactions of the Association for
  Computational Linguistics}}  \bibinfo{volume}{7} (\bibinfo{year}{2019}),
  \bibinfo{pages}{169--184}.
\newblock


\bibitem[\protect\citeauthoryear{Arora, Li, Liang, Ma, and Risteski}{Arora
  et~al\mbox{.}}{2016}]%
        {arora2016latent}
\bibfield{author}{\bibinfo{person}{Sanjeev Arora}, \bibinfo{person}{Yuanzhi
  Li}, \bibinfo{person}{Yingyu Liang}, \bibinfo{person}{Tengyu Ma}, {and}
  \bibinfo{person}{Andrej Risteski}.} \bibinfo{year}{2016}\natexlab{}.
\newblock \showarticletitle{A Latent Variable Model Approach to Pmi-Based Word
  Embeddings}.
\newblock \bibinfo{journal}{\emph{Transactions of the Association for
  Computational Linguistics}}  \bibinfo{volume}{4} (\bibinfo{year}{2016}),
  \bibinfo{pages}{385--399}.
\newblock


\bibitem[\protect\citeauthoryear{Barron}{Barron}{2019}]%
        {barron2019general}
\bibfield{author}{\bibinfo{person}{Jonathan~T Barron}.}
  \bibinfo{year}{2019}\natexlab{}.
\newblock \showarticletitle{A General and Adaptive Robust Loss Function}. In
  \bibinfo{booktitle}{\emph{Proceedings of the {{IEEE Conference}} on
  {{Computer Vision}} and {{Pattern Recognition}}}}.
  \bibinfo{pages}{4331--4339}.
\newblock


\bibitem[\protect\citeauthoryear{Beam, Kompa, Schmaltz, Fried, Weber, Palmer,
  Shi, Cai, and Kohane}{Beam et~al\mbox{.}}{2018}]%
        {beam2018clinical}
\bibfield{author}{\bibinfo{person}{Andrew~L Beam}, \bibinfo{person}{Benjamin
  Kompa}, \bibinfo{person}{Allen Schmaltz}, \bibinfo{person}{Inbar Fried},
  \bibinfo{person}{Griffin Weber}, \bibinfo{person}{Nathan Palmer},
  \bibinfo{person}{Xu Shi}, \bibinfo{person}{Tianxi Cai}, {and}
  \bibinfo{person}{Isaac~S Kohane}.} \bibinfo{year}{2018}\natexlab{}.
\newblock \showarticletitle{Clinical {{Concept Embeddings Learned}} from
  {{Massive Sources}} of {{Multimodal Medical Data}}}. In
  \bibinfo{booktitle}{\emph{Pacific {{Symposium}} on {{Biocomputing}}}},
  Vol.~\bibinfo{volume}{25}. \bibinfo{pages}{295--306}.
\newblock


\bibitem[\protect\citeauthoryear{Bojanowski, Grave, Joulin, and
  Mikolov}{Bojanowski et~al\mbox{.}}{2017}]%
        {bojanowski2017enriching}
\bibfield{author}{\bibinfo{person}{Piotr Bojanowski}, \bibinfo{person}{Edouard
  Grave}, \bibinfo{person}{Armand Joulin}, {and} \bibinfo{person}{Tomas
  Mikolov}.} \bibinfo{year}{2017}\natexlab{}.
\newblock \showarticletitle{Enriching Word Vectors with Subword Information}.
\newblock \bibinfo{journal}{\emph{Transactions of the Association for
  Computational Linguistics}}  \bibinfo{volume}{5} (\bibinfo{year}{2017}),
  \bibinfo{pages}{135--146}.
\newblock


\bibitem[\protect\citeauthoryear{Caruana}{Caruana}{1997}]%
        {caruana1997multitask}
\bibfield{author}{\bibinfo{person}{Rich Caruana}.}
  \bibinfo{year}{1997}\natexlab{}.
\newblock \showarticletitle{Multitask Learning}.
\newblock \bibinfo{journal}{\emph{Machine learning}} \bibinfo{volume}{28},
  \bibinfo{number}{1} (\bibinfo{year}{1997}), \bibinfo{pages}{41--75}.
\newblock


\bibitem[\protect\citeauthoryear{Cheng}{Cheng}{2004}]%
        {cheng2004study}
\bibfield{author}{\bibinfo{person}{Grace~YT Cheng}.}
  \bibinfo{year}{2004}\natexlab{}.
\newblock \showarticletitle{A Study of Clinical Questions Posed by Hospital
  Clinicians}.
\newblock \bibinfo{journal}{\emph{Journal of the Medical Library Association}}
  \bibinfo{volume}{92}, \bibinfo{number}{4} (\bibinfo{year}{2004}),
  \bibinfo{pages}{445}.
\newblock


\bibitem[\protect\citeauthoryear{Cohan, Dernoncourt, Kim, Bui, Kim, Chang, and
  Goharian}{Cohan et~al\mbox{.}}{2018}]%
        {cohan2018discourse}
\bibfield{author}{\bibinfo{person}{Arman Cohan}, \bibinfo{person}{Franck
  Dernoncourt}, \bibinfo{person}{Doo~Soon Kim}, \bibinfo{person}{Trung Bui},
  \bibinfo{person}{Seokhwan Kim}, \bibinfo{person}{Walter Chang}, {and}
  \bibinfo{person}{Nazli Goharian}.} \bibinfo{year}{2018}\natexlab{}.
\newblock \showarticletitle{A {{Discourse}}-{{Aware Attention Model}} for
  {{Abstractive Summarization}} of {{Long Documents}}}. In
  \bibinfo{booktitle}{\emph{Proceedings of the 2018 {{Conference}} of the
  {{North American Chapter}} of the {{Association}} for {{Computational
  Linguistics}}: {{Human Language Technologies}}}}. \bibinfo{pages}{615--621}.
\newblock


\bibitem[\protect\citeauthoryear{Dai, Xiong, Callan, and Liu}{Dai
  et~al\mbox{.}}{2018}]%
        {dai2018convolutional}
\bibfield{author}{\bibinfo{person}{Zhuyun Dai}, \bibinfo{person}{Chenyan
  Xiong}, \bibinfo{person}{Jamie Callan}, {and} \bibinfo{person}{Zhiyuan Liu}.}
  \bibinfo{year}{2018}\natexlab{}.
\newblock \showarticletitle{Convolutional Neural Networks for Soft-Matching
  n-Grams in Ad-Hoc Search}. In \bibinfo{booktitle}{\emph{Proceedings of the
  Eleventh {{ACM}} International Conference on Web Search and Data Mining}}.
  \bibinfo{publisher}{{ACM}}, \bibinfo{pages}{126--134}.
\newblock


\bibitem[\protect\citeauthoryear{Fujiwara, Yamamoto, Kim, Buske, and
  Takagi}{Fujiwara et~al\mbox{.}}{2018}]%
        {fujiwara2018pubcasefinder}
\bibfield{author}{\bibinfo{person}{Toyofumi Fujiwara},
  \bibinfo{person}{Yasunori Yamamoto}, \bibinfo{person}{Jin-Dong Kim},
  \bibinfo{person}{Orion Buske}, {and} \bibinfo{person}{Toshihisa Takagi}.}
  \bibinfo{year}{2018}\natexlab{}.
\newblock \showarticletitle{{{PubCaseFinder}}: {{A}} Case-Report-Based,
  Phenotype-Driven Differential-Diagnosis System for Rare Diseases}.
\newblock \bibinfo{journal}{\emph{The American Journal of Human Genetics}}
  \bibinfo{volume}{103}, \bibinfo{number}{3} (\bibinfo{year}{2018}),
  \bibinfo{pages}{389--399}.
\newblock


\bibitem[\protect\citeauthoryear{Gillick, Kulkarni, Lansing, Presta, Baldridge,
  Ie, and Garcia-Olano}{Gillick et~al\mbox{.}}{2019}]%
        {gillick2019learning}
\bibfield{author}{\bibinfo{person}{Daniel Gillick}, \bibinfo{person}{Sayali
  Kulkarni}, \bibinfo{person}{Larry Lansing}, \bibinfo{person}{Alessandro
  Presta}, \bibinfo{person}{Jason Baldridge}, \bibinfo{person}{Eugene Ie},
  {and} \bibinfo{person}{Diego Garcia-Olano}.} \bibinfo{year}{2019}\natexlab{}.
\newblock \showarticletitle{Learning {{Dense Representations}} for {{Entity
  Retrieval}}}. In \bibinfo{booktitle}{\emph{Proceedings of the 23rd
  {{Conference}} on {{Computational Natural Language Learning}} ({{CoNLL}})}}.
  \bibinfo{publisher}{{ACL}}, \bibinfo{pages}{528--537}.
\newblock


\bibitem[\protect\citeauthoryear{Gillick, Presta, and Tomar}{Gillick
  et~al\mbox{.}}{2018}]%
        {gillick2018end}
\bibfield{author}{\bibinfo{person}{Daniel Gillick}, \bibinfo{person}{Alessandro
  Presta}, {and} \bibinfo{person}{Gaurav~Singh Tomar}.}
  \bibinfo{year}{2018}\natexlab{}.
\newblock \showarticletitle{End-to-{{End Retrieval}} in {{Continuous Space}}}.
\newblock \bibinfo{journal}{\emph{arXiv:1811.08008 [cs.IR]}}
  (\bibinfo{year}{2018}).
\newblock


\bibitem[\protect\citeauthoryear{Gorman, Ash, and Wykoff}{Gorman
  et~al\mbox{.}}{1994}]%
        {gorman1994can}
\bibfield{author}{\bibinfo{person}{Paul~N Gorman}, \bibinfo{person}{Joan Ash},
  {and} \bibinfo{person}{Leslie Wykoff}.} \bibinfo{year}{1994}\natexlab{}.
\newblock \showarticletitle{Can Primary Care Physicians' Questions Be Answered
  Using the Medical Journal Literature?}
\newblock \bibinfo{journal}{\emph{Bulletin of the Medical Library Association}}
  \bibinfo{volume}{82}, \bibinfo{number}{2} (\bibinfo{year}{1994}),
  \bibinfo{pages}{140}.
\newblock


\bibitem[\protect\citeauthoryear{Guo, Fan, Ai, and Croft}{Guo
  et~al\mbox{.}}{2016}]%
        {guo2016deep}
\bibfield{author}{\bibinfo{person}{Jiafeng Guo}, \bibinfo{person}{Yixing Fan},
  \bibinfo{person}{Qingyao Ai}, {and} \bibinfo{person}{W~Bruce Croft}.}
  \bibinfo{year}{2016}\natexlab{}.
\newblock \showarticletitle{A Deep Relevance Matching Model for Ad-Hoc
  Retrieval}. In \bibinfo{booktitle}{\emph{Proceedings of the 25th {{ACM
  International Conference}} on {{Information}} and {{Knowledge Management}}}}.
  \bibinfo{publisher}{{ACM}}, \bibinfo{pages}{55--64}.
\newblock


\bibitem[\protect\citeauthoryear{Guo, Fan, Ji, and Cheng}{Guo
  et~al\mbox{.}}{2019}]%
        {guo2019matchzoo}
\bibfield{author}{\bibinfo{person}{Jiafeng Guo}, \bibinfo{person}{Yixing Fan},
  \bibinfo{person}{Xiang Ji}, {and} \bibinfo{person}{Xueqi Cheng}.}
  \bibinfo{year}{2019}\natexlab{}.
\newblock \showarticletitle{{{MatchZoo}}: {{A Learning}}, {{Practicing}}, and
  {{Developing System}} for {{Neural Text Matching}}}. In
  \bibinfo{booktitle}{\emph{Proceedings of the 42nd {{International ACM SIGIR
  Conference}} on {{Research}} and {{Development}} in {{Information
  Retrieval}}}}. \bibinfo{publisher}{{ACM}}, \bibinfo{pages}{1297--1300}.
\newblock


\bibitem[\protect\citeauthoryear{Hanauer, Mei, Law, Khanna, and Zheng}{Hanauer
  et~al\mbox{.}}{2015}]%
        {hanauer2015supporting}
\bibfield{author}{\bibinfo{person}{David~A Hanauer}, \bibinfo{person}{Qiaozhu
  Mei}, \bibinfo{person}{James Law}, \bibinfo{person}{Ritu Khanna}, {and}
  \bibinfo{person}{Kai Zheng}.} \bibinfo{year}{2015}\natexlab{}.
\newblock \showarticletitle{Supporting Information Retrieval from Electronic
  Health Records: {{A}} Report of {{University}} of {{Michigan}}'s Nine-Year
  Experience in Developing and Using the {{Electronic Medical Record Search
  Engine}} ({{EMERSE}})}.
\newblock \bibinfo{journal}{\emph{Journal of Biomedical Informatics}}
  \bibinfo{volume}{55} (\bibinfo{year}{2015}), \bibinfo{pages}{290--300}.
\newblock


\bibitem[\protect\citeauthoryear{Hochreiter and Schmidhuber}{Hochreiter and
  Schmidhuber}{1997}]%
        {hochreiter1997long}
\bibfield{author}{\bibinfo{person}{Sepp Hochreiter} {and}
  \bibinfo{person}{J{\"u}rgen Schmidhuber}.} \bibinfo{year}{1997}\natexlab{}.
\newblock \showarticletitle{Long Short-Term Memory}.
\newblock \bibinfo{journal}{\emph{Neural Computation}} \bibinfo{volume}{9},
  \bibinfo{number}{8} (\bibinfo{year}{1997}), \bibinfo{pages}{1735--1780}.
\newblock


\bibitem[\protect\citeauthoryear{Hu, Lu, Li, and Chen}{Hu
  et~al\mbox{.}}{2014}]%
        {hu2014convolutional}
\bibfield{author}{\bibinfo{person}{Baotian Hu}, \bibinfo{person}{Zhengdong Lu},
  \bibinfo{person}{Hang Li}, {and} \bibinfo{person}{Qingcai Chen}.}
  \bibinfo{year}{2014}\natexlab{}.
\newblock \showarticletitle{Convolutional Neural Network Architectures for
  Matching Natural Language Sentences}. In \bibinfo{booktitle}{\emph{Advances
  in Neural Information Processing Systems}}. \bibinfo{pages}{2042--2050}.
\newblock


\bibitem[\protect\citeauthoryear{Huang, He, Gao, Deng, Acero, and Heck}{Huang
  et~al\mbox{.}}{2013}]%
        {huang2013learning}
\bibfield{author}{\bibinfo{person}{Po-Sen Huang}, \bibinfo{person}{Xiaodong
  He}, \bibinfo{person}{Jianfeng Gao}, \bibinfo{person}{Li Deng},
  \bibinfo{person}{Alex Acero}, {and} \bibinfo{person}{Larry Heck}.}
  \bibinfo{year}{2013}\natexlab{}.
\newblock \showarticletitle{Learning Deep Structured Semantic Models for Web
  Search Using Clickthrough Data}. In \bibinfo{booktitle}{\emph{Proceedings of
  the 22nd {{ACM}} International Conference on {{Information}} \& {{Knowledge
  Management}}}}. \bibinfo{publisher}{{ACM}}, \bibinfo{pages}{2333--2338}.
\newblock


\bibitem[\protect\citeauthoryear{Huang, Lin, and Demner-Fushman}{Huang
  et~al\mbox{.}}{2006}]%
        {huang2006evaluation}
\bibfield{author}{\bibinfo{person}{Xiaoli Huang}, \bibinfo{person}{Jimmy Lin},
  {and} \bibinfo{person}{Dina Demner-Fushman}.}
  \bibinfo{year}{2006}\natexlab{}.
\newblock \showarticletitle{Evaluation of {{PICO}} as a Knowledge
  Representation for Clinical Questions}. In \bibinfo{booktitle}{\emph{{{AMIA}}
  Annual Symposium Proceedings}}, Vol.~\bibinfo{volume}{2006}.
  \bibinfo{publisher}{{AMIA}}, \bibinfo{pages}{359}.
\newblock


\bibitem[\protect\citeauthoryear{Huber}{Huber}{1992}]%
        {huber1992robust}
\bibfield{author}{\bibinfo{person}{Peter~J. Huber}.}
  \bibinfo{year}{1992}\natexlab{}.
\newblock \showarticletitle{Robust Estimation of a Location Parameter}.
\newblock In \bibinfo{booktitle}{\emph{Breakthroughs in Statistics}}.
  \bibinfo{publisher}{{Springer}}, \bibinfo{pages}{492--518}.
\newblock


\bibitem[\protect\citeauthoryear{Jin, Dhingra, Liu, Cohen, and Lu}{Jin
  et~al\mbox{.}}{2019}]%
        {jin2019pubmedqa}
\bibfield{author}{\bibinfo{person}{Qiao Jin}, \bibinfo{person}{Bhuwan Dhingra},
  \bibinfo{person}{Zhengping Liu}, \bibinfo{person}{William Cohen}, {and}
  \bibinfo{person}{Xinghua Lu}.} \bibinfo{year}{2019}\natexlab{}.
\newblock \showarticletitle{{{PubMedQA}}: {{A Dataset}} for {{Biomedical
  Research Question Answering}}}. In \bibinfo{booktitle}{\emph{Proceedings of
  the 2019 {{Conference}} on {{Empirical Methods}} in {{Natural Language
  Processing}} and the 9th {{International Joint Conference}} on {{Natural
  Language Processing}}}}. \bibinfo{pages}{2567--2577}.
\newblock


\bibitem[\protect\citeauthoryear{Jones}{Jones}{1972}]%
        {sparck1972statistical}
\bibfield{author}{\bibinfo{person}{Sparck~K. Jones}.}
  \bibinfo{year}{1972}\natexlab{}.
\newblock \showarticletitle{A {{Statistical}} Interpretation of Term
  Specificity and Its Application to Retrieval}.
\newblock \bibinfo{journal}{\emph{Journal of Documentation}}
  \bibinfo{volume}{28}, \bibinfo{number}{1} (\bibinfo{year}{1972}),
  \bibinfo{pages}{11--21}.
\newblock


\bibitem[\protect\citeauthoryear{Keikha, Park, Croft, and Sanderson}{Keikha
  et~al\mbox{.}}{2014}]%
        {keikha2014retrieving}
\bibfield{author}{\bibinfo{person}{Mostafa Keikha}, \bibinfo{person}{Jae~Hyun
  Park}, \bibinfo{person}{W.~Bruce Croft}, {and} \bibinfo{person}{Mark
  Sanderson}.} \bibinfo{year}{2014}\natexlab{}.
\newblock \showarticletitle{Retrieving Passages and Finding Answers}. In
  \bibinfo{booktitle}{\emph{Proceedings of the 2014 {{Australasian Document
  Computing Symposium}}}}. \bibinfo{publisher}{{ACM}}, \bibinfo{pages}{81}.
\newblock


\bibitem[\protect\citeauthoryear{Lee, Yoon, Kim, Kim, Kim, So, and Kang}{Lee
  et~al\mbox{.}}{2019}]%
        {lee2019biobert}
\bibfield{author}{\bibinfo{person}{Jinhyuk Lee}, \bibinfo{person}{Wonjin Yoon},
  \bibinfo{person}{Sungdong Kim}, \bibinfo{person}{Donghyeon Kim},
  \bibinfo{person}{Sunkyu Kim}, \bibinfo{person}{Chan~Ho So}, {and}
  \bibinfo{person}{Jaewoo Kang}.} \bibinfo{year}{2019}\natexlab{}.
\newblock \showarticletitle{{{BioBERT}}: A Pre-Trained Biomedical Language
  Representation Model for Biomedical Text Mining}.
\newblock \bibinfo{journal}{\emph{Bioinformatics}} (\bibinfo{year}{2019}),
  \bibinfo{pages}{1--7}.
\newblock


\bibitem[\protect\citeauthoryear{Logeswaran, Chang, Lee, Toutanova, Devlin, and
  Lee}{Logeswaran et~al\mbox{.}}{2019}]%
        {logeswaran2019zero}
\bibfield{author}{\bibinfo{person}{Lajanugen Logeswaran},
  \bibinfo{person}{Ming-Wei Chang}, \bibinfo{person}{Kenton Lee},
  \bibinfo{person}{Kristina Toutanova}, \bibinfo{person}{Jacob Devlin}, {and}
  \bibinfo{person}{Honglak Lee}.} \bibinfo{year}{2019}\natexlab{}.
\newblock \showarticletitle{Zero-{{Shot Entity Linking}} by {{Reading Entity
  Descriptions}}}. In \bibinfo{booktitle}{\emph{Proceedings of the 57th
  {{Annual Meeting}} of the {{Association}} for {{Computational
  Linguistics}}}}. \bibinfo{pages}{3449--3460}.
\newblock


\bibitem[\protect\citeauthoryear{Mitra, Diaz, and Craswell}{Mitra
  et~al\mbox{.}}{2017}]%
        {mitra2017learning}
\bibfield{author}{\bibinfo{person}{Bhaskar Mitra}, \bibinfo{person}{Fernando
  Diaz}, {and} \bibinfo{person}{Nick Craswell}.}
  \bibinfo{year}{2017}\natexlab{}.
\newblock \showarticletitle{Learning to Match Using Local and Distributed
  Representations of Text for Web Search}. In
  \bibinfo{booktitle}{\emph{Proceedings of the 26th {{International
  Conference}} on {{World Wide Web}}}}. \bibinfo{publisher}{{IW3C2}},
  \bibinfo{pages}{1291--1299}.
\newblock


\bibitem[\protect\citeauthoryear{Nanni, Mitra, Magnusson, and Dietz}{Nanni
  et~al\mbox{.}}{2017}]%
        {nanni2017benchmark}
\bibfield{author}{\bibinfo{person}{Federico Nanni}, \bibinfo{person}{Bhaskar
  Mitra}, \bibinfo{person}{Matt Magnusson}, {and} \bibinfo{person}{Laura
  Dietz}.} \bibinfo{year}{2017}\natexlab{}.
\newblock \showarticletitle{Benchmark for Complex Answer Retrieval}. In
  \bibinfo{booktitle}{\emph{Proceedings of the {{ACM SIGIR}} International
  Conference on Theory of Information Retrieval}}. \bibinfo{publisher}{{ACM}},
  \bibinfo{pages}{293--296}.
\newblock


\bibitem[\protect\citeauthoryear{Palangi, Deng, Shen, Gao, He, Chen, Song, and
  Ward}{Palangi et~al\mbox{.}}{2016}]%
        {palangi2016deep}
\bibfield{author}{\bibinfo{person}{Hamid Palangi}, \bibinfo{person}{Li Deng},
  \bibinfo{person}{Yelong Shen}, \bibinfo{person}{Jianfeng Gao},
  \bibinfo{person}{Xiaodong He}, \bibinfo{person}{Jianshu Chen},
  \bibinfo{person}{Xinying Song}, {and} \bibinfo{person}{Rabab Ward}.}
  \bibinfo{year}{2016}\natexlab{}.
\newblock \showarticletitle{Deep Sentence Embedding Using Long Short-Term
  Memory Networks: {{Analysis}} and Application to Information Retrieval}.
\newblock \bibinfo{journal}{\emph{IEEE/ACM Transactions on Audio, Speech and
  Language Processing (TASLP)}} \bibinfo{volume}{24}, \bibinfo{number}{4}
  (\bibinfo{year}{2016}), \bibinfo{pages}{694--707}.
\newblock


\bibitem[\protect\citeauthoryear{Palatucci, Pomerleau, Hinton, and
  Mitchell}{Palatucci et~al\mbox{.}}{2009}]%
        {palatucci2009zero}
\bibfield{author}{\bibinfo{person}{Mark Palatucci}, \bibinfo{person}{Dean
  Pomerleau}, \bibinfo{person}{Geoffrey~E Hinton}, {and} \bibinfo{person}{Tom~M
  Mitchell}.} \bibinfo{year}{2009}\natexlab{}.
\newblock \showarticletitle{Zero-Shot Learning with Semantic Output Codes}. In
  \bibinfo{booktitle}{\emph{Advances in Neural Information Processing
  Systems}}. \bibinfo{pages}{1410--1418}.
\newblock


\bibitem[\protect\citeauthoryear{Pang, Lan, Guo, Xu, Wan, and Cheng}{Pang
  et~al\mbox{.}}{2016}]%
        {pang2016text}
\bibfield{author}{\bibinfo{person}{Liang Pang}, \bibinfo{person}{Yanyan Lan},
  \bibinfo{person}{Jiafeng Guo}, \bibinfo{person}{Jun Xu},
  \bibinfo{person}{Shengxian Wan}, {and} \bibinfo{person}{Xueqi Cheng}.}
  \bibinfo{year}{2016}\natexlab{}.
\newblock \showarticletitle{Text Matching as Image Recognition}. In
  \bibinfo{booktitle}{\emph{Proceedings of the {{Thirtieth AAAI Conference}} on
  {{Artificial Intelligence}}}}. \bibinfo{pages}{2793--2799}.
\newblock


\bibitem[\protect\citeauthoryear{Pennington, Socher, and Manning}{Pennington
  et~al\mbox{.}}{2014}]%
        {pennington2014glove}
\bibfield{author}{\bibinfo{person}{Jeffrey Pennington},
  \bibinfo{person}{Richard Socher}, {and} \bibinfo{person}{Christopher~D.
  Manning}.} \bibinfo{year}{2014}\natexlab{}.
\newblock \showarticletitle{{{GloVe}}: {{Global Vectors}} for {{Word
  Representation}}}. In \bibinfo{booktitle}{\emph{Empirical {{Methods}} in
  {{Natural Language Processing}}}}. \bibinfo{pages}{1532--1543}.
\newblock


\bibitem[\protect\citeauthoryear{Richardson, Wilson, Nishikawa, Hayward, and
  {others}}{Richardson et~al\mbox{.}}{1995}]%
        {richardson1995well}
\bibfield{author}{\bibinfo{person}{W.~Scott Richardson},
  \bibinfo{person}{Mark~C. Wilson}, \bibinfo{person}{Jim Nishikawa},
  \bibinfo{person}{Robert~S. Hayward}, {and} \bibinfo{person}{{others}}.}
  \bibinfo{year}{1995}\natexlab{}.
\newblock \showarticletitle{The Well-Built Clinical Question: A Key to
  Evidence-Based Decisions}.
\newblock \bibinfo{journal}{\emph{ACP Journal Club}} \bibinfo{volume}{123},
  \bibinfo{number}{3} (\bibinfo{year}{1995}), \bibinfo{pages}{A12--3}.
\newblock


\bibitem[\protect\citeauthoryear{Robertson and Jones}{Robertson and
  Jones}{1976}]%
        {robertson1976relevance}
\bibfield{author}{\bibinfo{person}{Stephen~E. Robertson} {and}
  \bibinfo{person}{K.~Sparck Jones}.} \bibinfo{year}{1976}\natexlab{}.
\newblock \showarticletitle{Relevance Weighting of Search Terms}.
\newblock \bibinfo{journal}{\emph{Journal of the American Society for
  Information Science}} \bibinfo{volume}{27}, \bibinfo{number}{3}
  (\bibinfo{year}{1976}), \bibinfo{pages}{129--146}.
\newblock


\bibitem[\protect\citeauthoryear{Robertson, Walker, Jones, Hancock-Beaulieu,
  Gatford, and {others}}{Robertson et~al\mbox{.}}{1995}]%
        {robertson1995okapi}
\bibfield{author}{\bibinfo{person}{Stephen~E Robertson}, \bibinfo{person}{Steve
  Walker}, \bibinfo{person}{Susan Jones}, \bibinfo{person}{Micheline~M
  Hancock-Beaulieu}, \bibinfo{person}{Mike Gatford}, {and}
  \bibinfo{person}{{others}}.} \bibinfo{year}{1995}\natexlab{}.
\newblock \showarticletitle{Okapi at {{TREC}}-3}.
\newblock \bibinfo{journal}{\emph{NIST Special Publication SP}}
  \bibinfo{volume}{109} (\bibinfo{year}{1995}), \bibinfo{pages}{109}.
\newblock


\bibitem[\protect\citeauthoryear{Salton and Buckley}{Salton and
  Buckley}{1988}]%
        {salton1988term}
\bibfield{author}{\bibinfo{person}{Gerard Salton} {and}
  \bibinfo{person}{Christopher Buckley}.} \bibinfo{year}{1988}\natexlab{}.
\newblock \showarticletitle{Term-Weighting Approaches in Automatic Text
  Retrieval}.
\newblock \bibinfo{journal}{\emph{Information Processing \& Management}}
  \bibinfo{volume}{24}, \bibinfo{number}{5} (\bibinfo{year}{1988}),
  \bibinfo{pages}{513--523}.
\newblock


\bibitem[\protect\citeauthoryear{Schardt, Adams, Owens, Keitz, and
  Fontelo}{Schardt et~al\mbox{.}}{2007}]%
        {schardt2007utilization}
\bibfield{author}{\bibinfo{person}{Connie Schardt}, \bibinfo{person}{Martha~B.
  Adams}, \bibinfo{person}{Thomas Owens}, \bibinfo{person}{Sheri Keitz}, {and}
  \bibinfo{person}{Paul Fontelo}.} \bibinfo{year}{2007}\natexlab{}.
\newblock \showarticletitle{Utilization of the {{PICO}} Framework to Improve
  Searching {{PubMed}} for Clinical Questions}.
\newblock \bibinfo{journal}{\emph{BMC Medical Informatics and Decision Making}}
  \bibinfo{volume}{7}, \bibinfo{number}{1} (\bibinfo{year}{2007}),
  \bibinfo{pages}{16}.
\newblock


\bibitem[\protect\citeauthoryear{Schneider, Arnold, Oberhauser, Klatt, Steffek,
  and L{\"o}ser}{Schneider et~al\mbox{.}}{2018}]%
        {schneider2018smartmd}
\bibfield{author}{\bibinfo{person}{Rudolf Schneider},
  \bibinfo{person}{Sebastian Arnold}, \bibinfo{person}{Tom Oberhauser},
  \bibinfo{person}{Tobias Klatt}, \bibinfo{person}{Thomas Steffek}, {and}
  \bibinfo{person}{Alexander L{\"o}ser}.} \bibinfo{year}{2018}\natexlab{}.
\newblock \showarticletitle{Smart-{{MD}}: {{Neural Paragraph Retrieval}} of
  {{Medical Topics}}}. In \bibinfo{booktitle}{\emph{The {{Web Conference}} 2018
  {{Companion}}}}. \bibinfo{publisher}{{IW3C2}}, \bibinfo{pages}{203--206}.
\newblock


\bibitem[\protect\citeauthoryear{Seo, Kembhavi, Farhadi, and Hajishirzi}{Seo
  et~al\mbox{.}}{2017}]%
        {seo2017bidirectional}
\bibfield{author}{\bibinfo{person}{Minjoon Seo}, \bibinfo{person}{Aniruddha
  Kembhavi}, \bibinfo{person}{Ali Farhadi}, {and} \bibinfo{person}{Hannaneh
  Hajishirzi}.} \bibinfo{year}{2017}\natexlab{}.
\newblock \showarticletitle{Bidirectional Attention Flow for Machine
  Comprehension}.
\newblock \bibinfo{journal}{\emph{5th International Conference on Learning
  Representations}} (\bibinfo{year}{2017}).
\newblock


\bibitem[\protect\citeauthoryear{Serr{\`a} and Karatzoglou}{Serr{\`a} and
  Karatzoglou}{2017}]%
        {serra2017getting}
\bibfield{author}{\bibinfo{person}{Joan Serr{\`a}} {and}
  \bibinfo{person}{Alexandros Karatzoglou}.} \bibinfo{year}{2017}\natexlab{}.
\newblock \showarticletitle{Getting Deep Recommenders Fit: {{Bloom}} Embeddings
  for Sparse Binary Input/Output Networks}. In
  \bibinfo{booktitle}{\emph{Proceedings of the {{Eleventh ACM Conference}} on
  {{Recommender Systems}}}}. \bibinfo{publisher}{{ACM}},
  \bibinfo{pages}{279--287}.
\newblock


\bibitem[\protect\citeauthoryear{Shen, He, Gao, Deng, and Mesnil}{Shen
  et~al\mbox{.}}{2014}]%
        {shen2014learning}
\bibfield{author}{\bibinfo{person}{Yelong Shen}, \bibinfo{person}{Xiaodong He},
  \bibinfo{person}{Jianfeng Gao}, \bibinfo{person}{Li Deng}, {and}
  \bibinfo{person}{Gr{\'e}goire Mesnil}.} \bibinfo{year}{2014}\natexlab{}.
\newblock \showarticletitle{Learning Semantic Representations Using
  Convolutional Neural Networks for Web Search}. In
  \bibinfo{booktitle}{\emph{Proceedings of the 23rd {{International
  Conference}} on {{World Wide Web}}}}. \bibinfo{publisher}{{ACM}},
  \bibinfo{pages}{373--374}.
\newblock


\bibitem[\protect\citeauthoryear{Tellex, Katz, Lin, Fernandes, and
  Marton}{Tellex et~al\mbox{.}}{2003}]%
        {tellex2003quantitative}
\bibfield{author}{\bibinfo{person}{Stefanie Tellex}, \bibinfo{person}{Boris
  Katz}, \bibinfo{person}{Jimmy Lin}, \bibinfo{person}{Aaron Fernandes}, {and}
  \bibinfo{person}{Gregory Marton}.} \bibinfo{year}{2003}\natexlab{}.
\newblock \showarticletitle{Quantitative Evaluation of Passage Retrieval
  Algorithms for Question Answering}. In \bibinfo{booktitle}{\emph{Proceedings
  of the 26th Annual International {{ACM SIGIR}} Conference on {{Research}} and
  Development in Informaion Retrieval}}. \bibinfo{publisher}{{ACM}},
  \bibinfo{pages}{41--47}.
\newblock


\bibitem[\protect\citeauthoryear{Wan, Lan, Guo, Xu, Pang, and Cheng}{Wan
  et~al\mbox{.}}{2016}]%
        {wan2016deep}
\bibfield{author}{\bibinfo{person}{Shengxian Wan}, \bibinfo{person}{Yanyan
  Lan}, \bibinfo{person}{Jiafeng Guo}, \bibinfo{person}{Jun Xu},
  \bibinfo{person}{Liang Pang}, {and} \bibinfo{person}{Xueqi Cheng}.}
  \bibinfo{year}{2016}\natexlab{}.
\newblock \showarticletitle{A Deep Architecture for Semantic Matching with
  Multiple Positional Sentence Representations}. In
  \bibinfo{booktitle}{\emph{Proceedings of the {{Thirtieth AAAI Conference}} on
  {{Artificial Intelligence}}}}. \bibinfo{pages}{2835--2841}.
\newblock


\bibitem[\protect\citeauthoryear{Wang, Yang, Wei, Chang, and Zhou}{Wang
  et~al\mbox{.}}{2017}]%
        {wang2017gated}
\bibfield{author}{\bibinfo{person}{Wenhui Wang}, \bibinfo{person}{Nan Yang},
  \bibinfo{person}{Furu Wei}, \bibinfo{person}{Baobao Chang}, {and}
  \bibinfo{person}{Ming Zhou}.} \bibinfo{year}{2017}\natexlab{}.
\newblock \showarticletitle{Gated Self-Matching Networks for Reading
  Comprehension and Question Answering}. In
  \bibinfo{booktitle}{\emph{Proceedings of the 55th {{Annual Meeting}} of the
  {{Association}} for {{Computational Linguistics}}}}.
  \bibinfo{pages}{189--198}.
\newblock


\bibitem[\protect\citeauthoryear{Xiong, Dai, Callan, Liu, and Power}{Xiong
  et~al\mbox{.}}{2017}]%
        {xiong2017end}
\bibfield{author}{\bibinfo{person}{Chenyan Xiong}, \bibinfo{person}{Zhuyun
  Dai}, \bibinfo{person}{Jamie Callan}, \bibinfo{person}{Zhiyuan Liu}, {and}
  \bibinfo{person}{Russell Power}.} \bibinfo{year}{2017}\natexlab{}.
\newblock \showarticletitle{End-to-End Neural Ad-Hoc Ranking with Kernel
  Pooling}. In \bibinfo{booktitle}{\emph{Proceedings of the 40th
  {{International ACM SIGIR}} Conference on Research and Development in
  Information Retrieval}}. \bibinfo{publisher}{{ACM}}, \bibinfo{pages}{55--64}.
\newblock


\bibitem[\protect\citeauthoryear{Yang, Ai, Guo, and Croft}{Yang
  et~al\mbox{.}}{2016a}]%
        {yang2016anmm}
\bibfield{author}{\bibinfo{person}{Liu Yang}, \bibinfo{person}{Qingyao Ai},
  \bibinfo{person}{Jiafeng Guo}, {and} \bibinfo{person}{W~Bruce Croft}.}
  \bibinfo{year}{2016}\natexlab{a}.
\newblock \showarticletitle{{{aNMM}}: {{Ranking}} Short Answer Texts with
  Attention-Based Neural Matching Model}. In
  \bibinfo{booktitle}{\emph{Proceedings of the 25th {{ACM}} International on
  Conference on Information and Knowledge Management}}.
  \bibinfo{publisher}{{ACM}}, \bibinfo{pages}{287--296}.
\newblock


\bibitem[\protect\citeauthoryear{Yang, Ai, Spina, Chen, Pang, Croft, Guo, and
  Scholer}{Yang et~al\mbox{.}}{2016b}]%
        {yang2016beyond}
\bibfield{author}{\bibinfo{person}{Liu Yang}, \bibinfo{person}{Qingyao Ai},
  \bibinfo{person}{Damiano Spina}, \bibinfo{person}{Ruey-Cheng Chen},
  \bibinfo{person}{Liang Pang}, \bibinfo{person}{W~Bruce Croft},
  \bibinfo{person}{Jiafeng Guo}, {and} \bibinfo{person}{Falk Scholer}.}
  \bibinfo{year}{2016}\natexlab{b}.
\newblock \showarticletitle{Beyond Factoid {{QA}}: Effective Methods for
  Non-Factoid Answer Sentence Retrieval}. In \bibinfo{booktitle}{\emph{European
  {{Conference}} on {{Information Retrieval}}}}.
  \bibinfo{publisher}{{Springer}}, \bibinfo{pages}{115--128}.
\newblock


\bibitem[\protect\citeauthoryear{Zhang and Zhou}{Zhang and Zhou}{2006}]%
        {zhang2006multilabel}
\bibfield{author}{\bibinfo{person}{Min-Ling Zhang} {and}
  \bibinfo{person}{Zhi-Hua Zhou}.} \bibinfo{year}{2006}\natexlab{}.
\newblock \showarticletitle{Multilabel Neural Networks with Applications to
  Functional Genomics and Text Categorization}.
\newblock \bibinfo{journal}{\emph{IEEE transactions on Knowledge and Data
  Engineering}} \bibinfo{volume}{18}, \bibinfo{number}{10}
  (\bibinfo{year}{2006}), \bibinfo{pages}{1338--1351}.
\newblock


\bibitem[\protect\citeauthoryear{Zhu, Ahuja, Wei, and Reddy}{Zhu
  et~al\mbox{.}}{2019}]%
        {zhu2019hierarchical}
\bibfield{author}{\bibinfo{person}{Ming Zhu}, \bibinfo{person}{Aman Ahuja},
  \bibinfo{person}{Wei Wei}, {and} \bibinfo{person}{Chandan~K. Reddy}.}
  \bibinfo{year}{2019}\natexlab{}.
\newblock \showarticletitle{A {{Hierarchical Attention Retrieval Model}} for
  {{Healthcare Question Answering}}}. In \bibinfo{booktitle}{\emph{The {{World
  Wide Web Conference}}}}. \bibinfo{publisher}{{ACM}},
  \bibinfo{pages}{2472--2482}.
\newblock


\end{thebibliography}

\end{document}